\documentclass[10pt,twocolumn,letterpaper]{article}
\pdfoutput=1

\usepackage{titling} % title in supp

\usepackage{cvpr}
\usepackage{times}
\usepackage{epsfig}
\usepackage{graphicx}
\usepackage{amsmath}
\usepackage{amssymb}

\usepackage{multirow}
\usepackage{enumitem}
\usepackage{algorithm2e}
\usepackage{color,xcolor,colortbl}
\usepackage{textcomp}
\usepackage{array}

\usepackage{booktabs}
\usepackage{float}
\usepackage{caption}
\usepackage{comment}
\usepackage{subcaption}
\usepackage[export]{adjustbox}
\usepackage{authblk}

\newcommand{\chk}{\checkmark}
\newcommand{\X}{$\times$}

\newcommand{\apprx}{\raise.17ex\hbox{$\scriptstyle\sim$}}

\usepackage[pagebackref=true,breaklinks=true,letterpaper=true,colorlinks,bookmarks=false]{hyperref}

\newcommand\customparagraph[1]{\vspace{0.6em}\noindent\textbf{#1}}

\cvprfinalcopy % *** Uncomment this line for the final submission

\def\cvprPaperID{2286} % *** Enter the CVPR Paper ID here
\def\httilde{\mbox{\tt\raisebox{-.5ex}{\symbol{126}}}}

\ifcvprfinal\fi

\begin{document}

\pagenumbering{gobble}

\title{\vspace*{-2.5ex} Revealing Scenes by Inverting Structure from Motion Reconstructions \vspace*{-1.5ex}}

\author{Francesco Pittaluga$^{1}$ \quad Sanjeev J.~Koppal$^{1}$ \quad Sing Bing Kang$^{2}$ \quad Sudipta N.~Sinha$^{2}$\\
$^1$ University of Florida \qquad $^2$ Microsoft Research
}

\twocolumn[{%
\renewcommand\twocolumn[1][]{#1}%
\maketitle
\vspace{-10mm}
\begin{center}
\fbox{\includegraphics[scale=0.288]{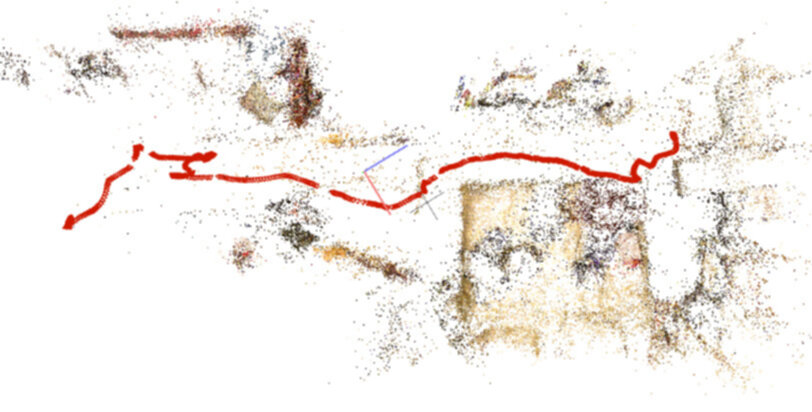}}
\fbox{\includegraphics[scale=0.18]{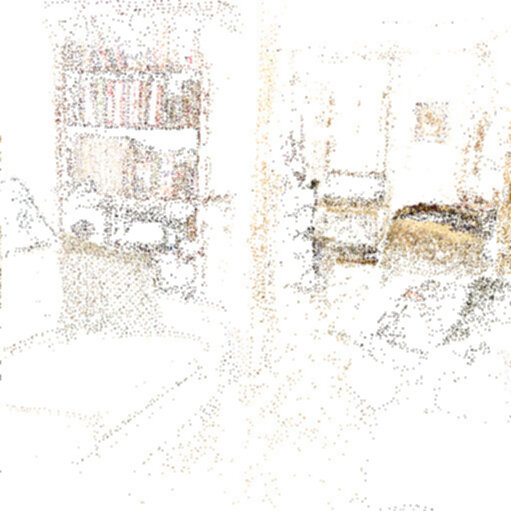}}
\fbox{\includegraphics[scale=0.18]{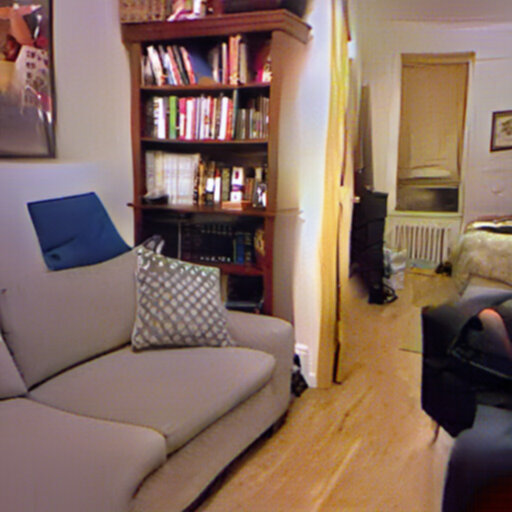}}
\fbox{\includegraphics[scale=0.18]{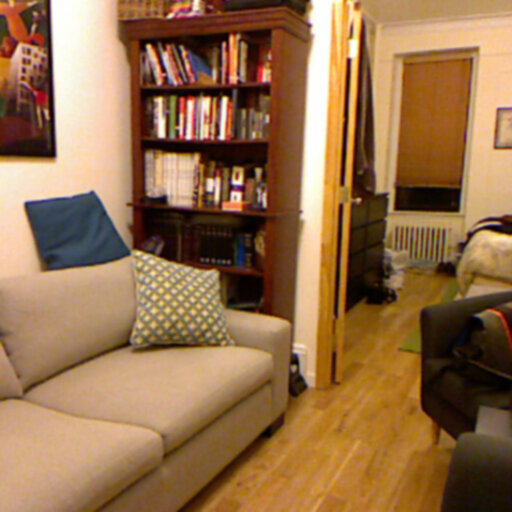}}\\
\vspace{1mm}
\hspace{9mm} (a) SfM point cloud (top view) \hspace{9mm} (b) Projected 3D points \hspace{2mm} (c) Synthesized Image \hspace{4mm} (d) Original Image
\captionof{figure}{\textsc{Synthesizing imagery from a SfM point cloud:}
From left to right:
(a) Top view of a SfM reconstruction of an indoor scene,
(b) 3D points projected into a viewpoint associated with a source image, (c) the image reconstructed using our technique, and (d) the source image. The reconstructed image is very detailed and closely resembles the source image.}
\label{teaser}
\end{center}
}]

\begin{abstract}
Many 3D vision systems localize cameras within a scene using 3D point clouds. Such point clouds are often obtained using structure from motion (SfM), after which the images are discarded to preserve privacy. In this paper, we show, for the first time, that such point clouds retain enough information to reveal scene appearance and compromise privacy.
We present a privacy attack that reconstructs color images of the scene from the point cloud. Our method is based on a cascaded U-Net that takes as input, a 2D multichannel image of the points rendered from a specific viewpoint containing point depth and optionally color and SIFT descriptors and outputs a color image of the scene from that viewpoint.
Unlike previous feature inversion methods~\cite{weinzaepfel2011reconstructing, dosovitskiy2016inverting}, we deal with highly sparse and irregular 2D point distributions and inputs where many point attributes are missing, namely keypoint orientation and scale, the descriptor image source and the 3D point visibility.
We evaluate our attack algorithm on public datasets~\cite{MegaDepth,NYU} and analyze the significance of the point cloud attributes. 
Finally, we show that novel views can also be generated thereby enabling compelling virtual tours of the underlying scene.
\end{abstract} 
\vspace{-1.5em}
\section{Introduction}

Emerging AR technologies on mobile devices based on ARCore~\cite{ARCore}, ARKit~\cite{ARKit},
3D mapping APIs~\cite{6Dai}, and new devices such as HoloLens~\cite{Hololens} have set
the stage for deployment of devices with always-on cameras in our homes, workplaces, and other sensitive environments. Image-based localization techniques allow such devices to estimate their precise pose
within the scene~\cite{irschara2009,sattler2011,li2012,lim2015}. However, these localization methods requires persistent storage of 3D models of the
scene which contains sparse 3D point clouds reconstructed using images and SfM algorithms~\cite{schoenberger2016sfm}.

SfM source images are usually discarded to safeguard privacy. Surprisingly, however, we show that the SfM point
cloud and the associated attributes such as color and SIFT descriptors contain enough information
to reconstruct detailed comprehensible images of the scene (see Fig.~\ref{teaser} and Fig.~\ref{fig_md_nyu}).
This suggests that the persistent point cloud storage poses serious privacy risks that have been
widely ignored so far but will become increasingly relevant as localization services are adopted by a
larger user community.

While privacy issues for wearable devices have been studied~\cite{hong2013considering},
to the best of our knowledge, a systematic analysis of privacy risk of storing 3D point cloud maps has never been reported. We illustrate the privacy concerns by proposing the problem of synthesizing color images from an SfM model of a scene. We assume that the reconstructed model contains a sparse 3D point cloud with optional attributes such as descriptors, color, point visibility and associated camera poses but not the source images.

We make the following contributions: \textbf{(1)}~We introduce the problem of inverting a sparse SfM point cloud and reconstructing detailed views of the scene from arbitrary viewpoints. This problem differs from the previously studied single-image feature inversion problem due to the need to deal with highly sparse point distributions and a higher degree of missing information in the input, namely unknown keypoint orientation and scale, unknown image source of descriptors, and unknown 3D point visibilities.
\textbf{(2)}~We present a new approach based on three neural networks where the first network performs visibility estimation, the second network reconstructs the image and the third network uses an adversarial framework to further refine the image quality.
\textbf{(3)}~We systematically analyze variants of the inversion attack that exploits additional attributes that may be available, namely per-point descriptors, color and information about the source camera poses and point visibility and show that even the minimalist representation (descriptors only) are prone to the attack.
\textbf{(4)}~
We demonstrate the need for developing privacy preserving 3D representations, since the reconstructed images reveal the scene in great details and confirm the feasibility of the attack in a wide range of scenes. We also show that novel views of the scene can be synthesized
without any additional effort and a compelling virtual tour of a scene can be easily generated.

The three networks in our cascade are trained on 700+ indoor and outdoor SfM reconstructions generated from 500k+ multi-view images taken from the NYU2 \cite{NYU} and MegaDepth \cite{MegaDepth} datasets. The training data for all three networks including the visibility labels were generated automatically using COLMAP~\cite{schoenberger2016sfm}.
Next we compare our approach to previous work on inverting image features ~\cite{weinzaepfel2011reconstructing,dosovitskiy2016inverting,dosovitskiy2016generating} and discuss how the problem of inverting SfM models poses a unique set of challenges.

\section{Related Work}

In this section, we review existing work on inverting image features
and contrast them to inverting SfM point cloud
models. We then broadly discuss image-to-image translation, upsampling and interpolation, and privacy attacks.

\customparagraph{Inverting features.} The task of reconstructing images from features has been explored to understand what is encoded by the features, as was done for SIFT features by Weinzaepfel et al. \cite{weinzaepfel2011reconstructing}, HOG features by Vondrick et al.~\cite{vondrick2013hoggles} and bag-of-words by Kato and Harada~\cite{kato2014image}. Recent work on the topic has been primarily focused on inverting and interpreting CNN features~\cite{zeiler2014visualizing,yosinski2015understanding,mahendran2015understanding}. Dosovitskiy and Brox proposed encoder-decoder CNN architectures for
inverting many different features (DB1)~\cite{dosovitskiy2016inverting} and later incorporated adversarial training with perceptual loss functions (DB2)~\cite{dosovitskiy2016generating}.
While DB1~\cite{dosovitskiy2016inverting} showed some qualitative results on inverting sparse SIFT, both papers focused primarily on dense features.
In contrast to these feature inversion approaches, we focus solely on inverting SIFT descriptors stored along with SfM point clouds. While the projected 3D points on a chosen viewpoint may resemble single image SIFT features, there are some key differences. First, our input 2D point distributions can be highly sparse and irregular, due to the typical inherent sparsity of SfM point clouds. Second, the SIFT keypoint scale and orientation are unknown since SfM methods retain only the descriptors for the 3D points. Third, each 3D point typically has only one descriptor sampled from
an arbitrary source image whose identity is not stored either, entailing descriptors with unknown perspective distortions and photometric inconsistencies. Finally, the 3D point visibilities are also unknown and we will demonstrate the importance of visibility reasoning in the paper.

\customparagraph{Image-to-Image Translation.}
Various methods such as Pix2Pix \cite{isola2017image}, CycleGan \cite{zhu2017unpaired}, CoGAN \cite{liu2016coupled}
and related unsupervised approaches \cite{donahue2016adversarial,liu2017unsupervised,qi2018}
use conditional adversarial networks to transform between 2D representations, such as edge to color, label to color, and day to night images. While such networks are typically dense (without holes) and usually low-dimensional (single channel or RGB), Contour2Im~\cite{dekel2017smart} takes sparse 2D points sampled along gradients along with low-dimensional input features. In contrast to our work, these approaches are trained on specific object categories and semantically similar images. While we use similar building blocks to these methods (encoder-decoder networks, U-nets, adversarial loss, and perceptual loss), our networks can generalize to arbitrary images, and are trained on large scale indoor and outdoor SfM datasets.

\customparagraph{Upsampling.} When the input and output domains are identical, deep networks have shown excellent results on upsampling and superresolution tasks for images, disparity, depth maps and active range maps~\cite{chen2018estimating,lu2015sparse,uhrig2017sparsity,riegler2016atgv,hui16msgnet}. However, prior upsampling methods typically focus on inputs with uniform sparsity. Our approach differs due to the non-uniform spatial sampling in the input data which also happens to be high dimensional and noisy since the input descriptors are from different source images and viewpoints.

\customparagraph{Novel view synthesis and image-based rendering.}
Deep networks can significantly improve photorealism in free viewpoint image-based rendering~\cite{deepstereo,Hedman2018}. Additionally, several works have also explored monocular depth estimation and novel view synthesis using U-Nets \cite{eigen2014depth,MegaDepth,moukari2018deep}. Our approach arguably provides similar photorealistic visually quality -- remarkably, from sparse SfM reconstructions instead of images. This is disappointing news from a privacy perspective but could be useful in other settings for generating photorealistic images from 3D reconstructions.

\customparagraph{CNN-based privacy attacks and defense techniques.}
Recently, McPherson et al.~\cite{mcpherson2016} and Vasiljevic et al.~\cite{vasiljevic2016examining} showed that deep models could defeat existing image obfuscation methods. Further more, many image transformations can be considered as adding noise and undoing them as denoising, and here deep networks have been quite successful~\cite{xu2014deep}. To defend against CNN-based attacks, attempts at learning CNN-resistant transformations have shown some promise~\cite{pittaluga2019learning,edwards2015censoring,raval2017protecting,hamm2017minimax}. Concurrent to our work, Speciale et al.~\cite{speciale2019} introduced the privacy preserving image-based localization problem to address the privacy issues we have brought up. They proposed a new camera pose estimation technique using an obfuscated representation of the map geometry which can defend against our inversion attack.

\begin{figure*}
\centering
\includegraphics[width=\linewidth]{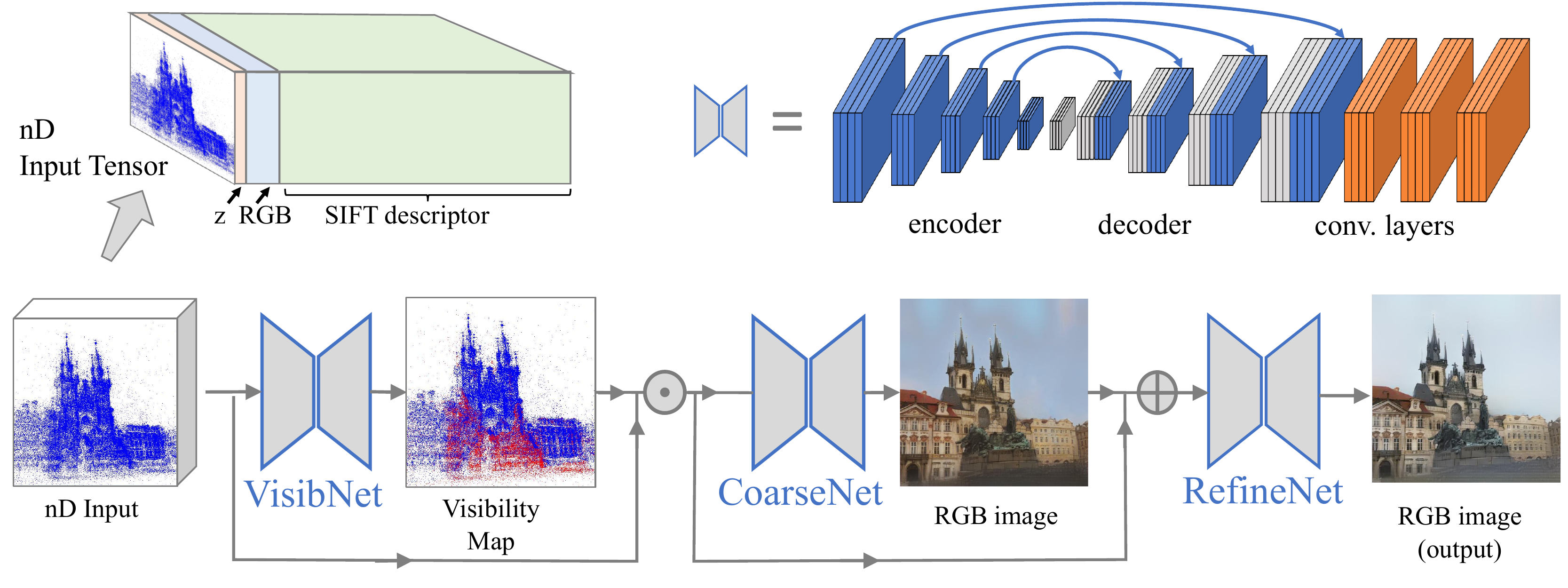}
\vspace{-.5cm}
\caption{\textsc{Network Architecture:} Our network has three sub-networks -- \textsc{VisibNet}, \textsc{CoarseNet} and \textsc{RefineNet}. The upper left shows that the input to our network is a multi-dimensional nD array. The paper explores network variants where the inputs are different subsets of depth, color and SIFT descriptors.
The three sub-networks have similar architectures. They are U-Nets with encoder and decoder layers with symmetric skip connections. The extra layers at the end of the decoder layers (marked in orange) are there to help with high-dimensional inputs. See the text and supplementary material for details.}
\vspace{-.2cm}
\label{fig:netarch}
\end{figure*}

\section{Method}

The input to our pipeline is a feature map generated from a SfM 3D point cloud model given a specific viewpoint \ie a set of camera extrinsic parameters. We obtain this feature map by projecting the 3D points on the image plane and associating the 3D point attributes (SIFT descriptor, color, etc.) with the discrete 2D pixel where the 3D point projects in the image. When multiple points project to the same pixel, we retain the attributes for the point closest to the camera and store its depth. We train a cascade of three encoder-decoder neural networks for visibility estimation, coarse image reconstruction and the final refinement step which recovers fine details in the reconstructed image.

\customparagraph{Visibility Estimation.} Since SfM 3D point clouds are often quite sparse and the underlying geometry and topology of the surfaces in the scene are unknown, it is not possible to easily determine which 3D points should be considered as visible from a specific camera viewpoint just using z-buffering. This is because a sufficient number of 3D points may not have been reconstructed on the foreground occluding surfaces.
This produces 2D pixels in the input feature maps which are associated with 3D points in the background \ie lie on surfaces which are occluded from that viewpoint.
Identifying and removing such points from the feature maps is critical to generating high-quality images and avoiding visual artifacts. We propose to recover point visibility using a data-driven neural network-based approach, which we refer to as \textsc{VisibNet}. We also evaluate two geometric methods which we refer to as \textsc{VisibSparse} and \textsc{VisibDense}. Both geometric methods however require additional information which might be unavailable.

\customparagraph{Coarse Image Reconstructon and Refinement.} Our technique for image synthesis from feature maps consists of a coarse image reconstruction step followed by a refinement step. \textsc{CoarseNet} is conditioned on the input feature map and produces an RGB image of the same width and height as the feature map. \textsc{RefineNet} outputs the final color image which has the same size, given the input feature map along with the image output of \textsc{CoarseNet} as its input.

\subsection{Visibility Estimation}
\label{ssec:visibest}

If we did not perform explicit visibility prediction in our pipeline, some degree of implicit visibility reasoning would still be carried out by the image synthesis network \textsc{CoarseNet}. In theory, this network has access to the input depths and could learn to reason about visibility. However, in practice, we found that this approach to be inaccurate, especially in regions where the input feature maps contain a low ratio of visible to occluded points. Qualitative examples of these failure cases are shown in Figure \ref{fig:visibcmp}. Therefore we explored explicit visibility estimation approaches based on geometric reasoning as well as learning.

\customparagraph{VisibSparse.} 
We explored a simple geometric method that we refer to as \textsc{VisibSparse}. It is based on the ``point splatting" paradigm used in computer graphics. By considering only the depth channel in the input, we apply a $\min$ filter with a $k \times k$ kernel on the feature map to obtain a filtered depth map. Here, we used $k=3$ based on empirical testing. Each entry in the feature map whose depth value is no greater than $5\%$ of the depth value in the filtered depth map is retained as visible. Otherwise, the point is considered occluded and the associated entry in the input is removed.

\customparagraph{VisibDense.}
When the camera poses for the source images computed during SfM and the image measurements are stored along with the 3D point cloud, it is often possible to exploit that data to compute a dense scene reconstruction. Labatut et al.~\cite{labatut2007} proposed such a method to compute a dense triangulated mesh by running space carving on the tetrahedral cells of the 3D Delaunay triangulation of the sparse SfM points. We used this method, implemented in COLMAP~\cite{schoenberger2016sfm} and 
computed 3D point visibility based on the reconstructed mesh model using traditional z-buffering.

\customparagraph{VisibNet.} A geometric method such as \textsc{VisibDense} cannot be used when the SfM cameras poses and image measurements are unavailable. We therefore propose a general regression-based approach that directly predicts the visibility from the input feature maps, where the predictive model is trained using supervised learning. Specifically, we train an encoder-decoder neural network which we refer to as \textsc{VisibNet} to classify each input point as either ``visible'' or ``occluded''. Ground truth visibility labels were generated automatically by leveraging \textsc{VisibDense} on all train, test, and validation scenes. Using \textsc{VisibNet}'s predictions to ``cull'' occluded points from the input feature maps prior to running \textsc{CoarseNet} significantly improves the quality of the reconstructed images, especially in regions where the input feature map contains fewer visible points compared to the number of points that are actually occluded.

\subsection{Architecture}

A sample input feature map as well as our complete network architecture consisting of \textsc{VisibNet}, \textsc{CoarseNet}, and \textsc{RefineNet} is shown in Figure~\ref{fig:netarch}. The input to our network is an $H\times W\times n$ dimensional feature map consisting of $n$-dimensional feature vectors with different combinations of depth, color, and SIFT features at each 2D location. Except for the number of input/output channels in the first/final layers, each sub-network has the same architecture consisting of U-Nets with a series of encoder-decoder layers with skip connections. Compared to conventional U-Nets, our network has a few extra convolutional layers at the end of the decoder layers. These extra layers facilitate propagation of information from the low-level features, particularly the information extracted from SIFT descriptors, via the skip connections to a larger pixel area in the output, while also helping to attenuate visual artifacts resulting from the highly sparse and irregular distribution of these features. We use nearest neighbor upsampling followed by standard convolutions instead of transposed convolutions as the latter are known to produce artifacts~\cite{odena2016deconvolution}.

\subsection{Optimization}

We separately train the sub-networks in our architecture, \textsc{VisibNet}, \textsc{CoarseNet}, and \textsc{RefineNet}. Batch normalization was used in every layer, except the final one in each network. We applied Xavier initialization and projections were generated on-the-fly to facilitate data augmentation during training and novel view generation after training.
\textsc{VisibNet} was trained first to classify feature map points as either visible or occluded, using ground-truth visibility masks generated automatically by running \textsc{VisibDense} for all train, test, and validation samples. Given training pairs of input feature maps $F_{x}\in \mathbb{R}^{H\times W\times N}$ and target source images $x \in \mathbb{R}^{H\times W\times 3}$, \textsc{VisibNet}'s objective is
\begin{equation}
\begin{split}
\mathcal{L}_{V}(x) = - \sum_{i=1}^{M} \big[ & U_x\text{log}\big((V(F_x)+1)/2\big) + \\
                                      & (1-U_x)\text{log}\big((1-V(F_x))/2\big) \big]_i,
\end{split}
\label{eq:ce}
\end{equation}
where $V: \mathbb{R}^{H\times W\times N} \rightarrow \mathbb{R}^{W\times H\times 1}$ denotes a differentiable function representing \textsc{VisibNet}, with learnable parameters, $U_x \in \mathbb{R}^{H\times W\times 1}$ denotes the ground-truth visibility map for feature map $F_x$, and the summation is carried out over the set of $M$ non-zero spatial locations in $F_x$.

\textsc{CoarseNet} was trained next, using a combination of an L1 pixel loss and an L2 perceptual loss (as in \cite{ledig2017photo,  dosovitskiy2016generating}) over the outputs of layers \emph{relu1\_1}, \emph{relu2\_2}, and \emph{relu3\_3} of VGG16 \cite{simonyan2014very} pre-trained for image classification on the ImageNet \cite{deng2009imagenet} dataset. The weights of \textsc{VisibNet} remained fixed while \textsc{CoarseNet} was being trained using the loss
\begin{equation}
\mathcal{L}_{C} = ||C(F_x)-x||_1 + \alpha \sum_{i=1}^3 ||\phi_i(C(F_x))-\phi_i(x)||_2^2,
\end{equation}
where $C: \mathbb{R}^{H\times W\times N} \rightarrow \mathbb{R}^{H\times W\times 3}$ denotes a differentiable function representing \textsc{CoarseNet}, with learnable parameters, and
$\phi_1: \mathbb{R}^{H\times W\times 3} \rightarrow \mathbb{R}^{\frac{H}{2}\times \frac{W}{2}\times 64}$,
$\phi_2: \mathbb{R}^{H\times W\times 3} \rightarrow \mathbb{R}^{\frac{H}{4}\times \frac{W}{4}\times 128}$, and
$\phi_3: \mathbb{R}^{H\times W\times 3} \rightarrow \mathbb{R}^{\frac{H}{8}\times \frac{W}{8}\times 256}$ denote the layers \emph{relu1\_1}, \emph{relu2\_2}, and \emph{relu2\_2}, respectively, of the pre-trained VGG16 network.

\textsc{RefineNet} was trained last using a combination of an L1 pixel loss, the same L2 perceptual loss as \textsc{CoarseNet}, and an adversarial loss. While training \textsc{RefineNet}, the weights of \textsc{VisibNet} and \textsc{CoarseNet} remained fixed. For adversarial training, we used a conditional discriminator whose goal was to distinguish between real source images used to generate the SfM models and images synthesized by \textsc{RefineNet}. The discriminator trained using cross-entropy loss similar to Eq.~(\ref{eq:ce}). Additionally, to stabilize adversarial training, $\phi_1(R(F_x))_{1}$, $\phi_2(R(F_x))_{1}$, and $\phi_3(R(F_x))_{1}$ were concatenated before the first, second, and third convolutional layers of the discriminator as done in \cite{sungatullina2018image}. \textsc{RefineNet} denoted as $R()$ has the following loss.
\begin{equation}
\begin{split}
\mathcal{L}_{R} = & ||R(F_x)-x||_1 + \alpha \sum_{i=1}^3 ||\phi_i(R(F_x))-\phi_i(x)||_2^2 \\
                  & + \beta[\text{log}(D(x)) + \text{log}(1-D(R(F_x)))].
\end{split}
\end{equation}
Here, the two functions, $R:\mathbb{R}^{H\times W\times N+3}\,\rightarrow\,\mathbb{R}^{H\times W\times 3}$ and
$D: \mathbb{R}^{H\times W\times N+3} \rightarrow \mathbb{R}$ denote differentiable functions representing \textsc{RefineNet} and the discriminator, respectively, with learnable parameters. We trained \textsc{RefineNet} to minimize $\mathcal{L}_{R}$ by applying alternating gradient updates to \textsc{RefineNet} and the discriminator. The gradients were computed on mini-batches of training data, with different batches used to update \textsc{RefineNet} and the discriminator.
\begin{figure*}
\centering
\includegraphics[width=0.12\linewidth]{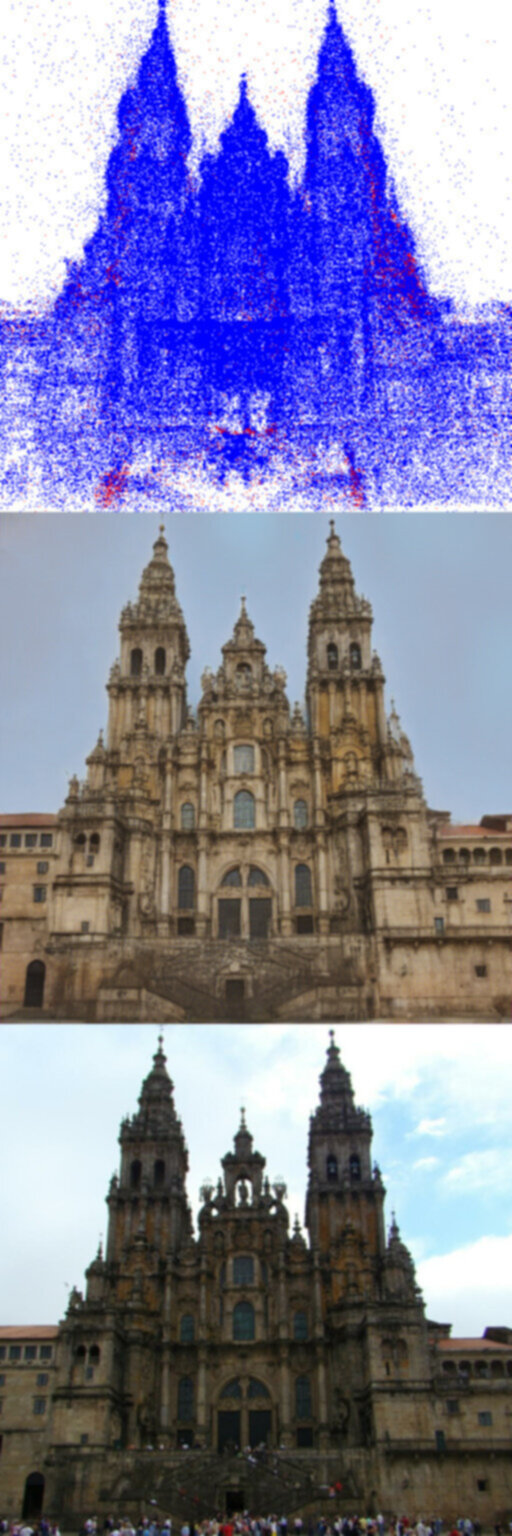}
\includegraphics[width=0.12\linewidth]{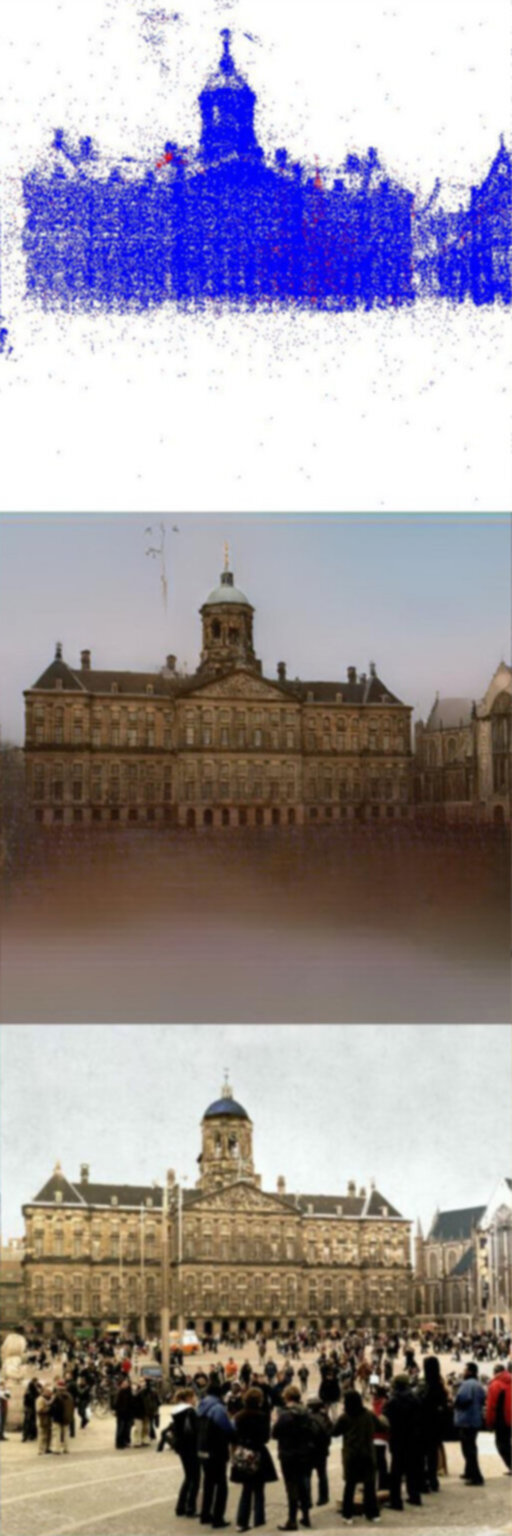}
\includegraphics[width=0.12\linewidth]{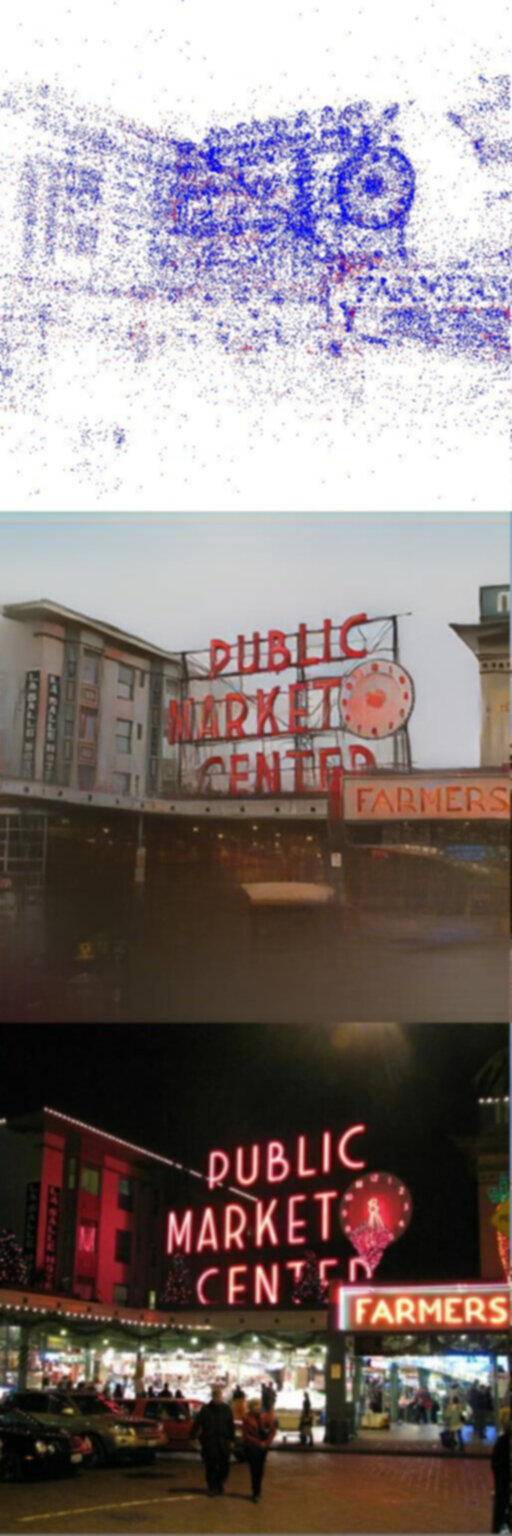}
\includegraphics[width=0.12\linewidth]{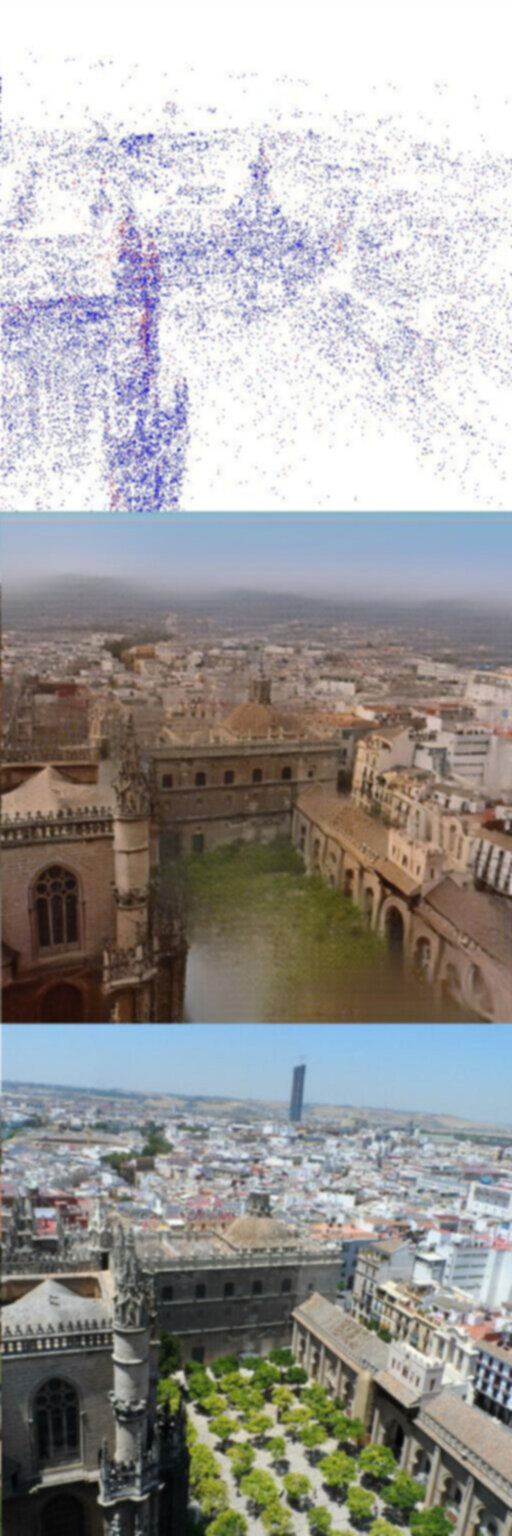}
\includegraphics[width=0.12\linewidth]{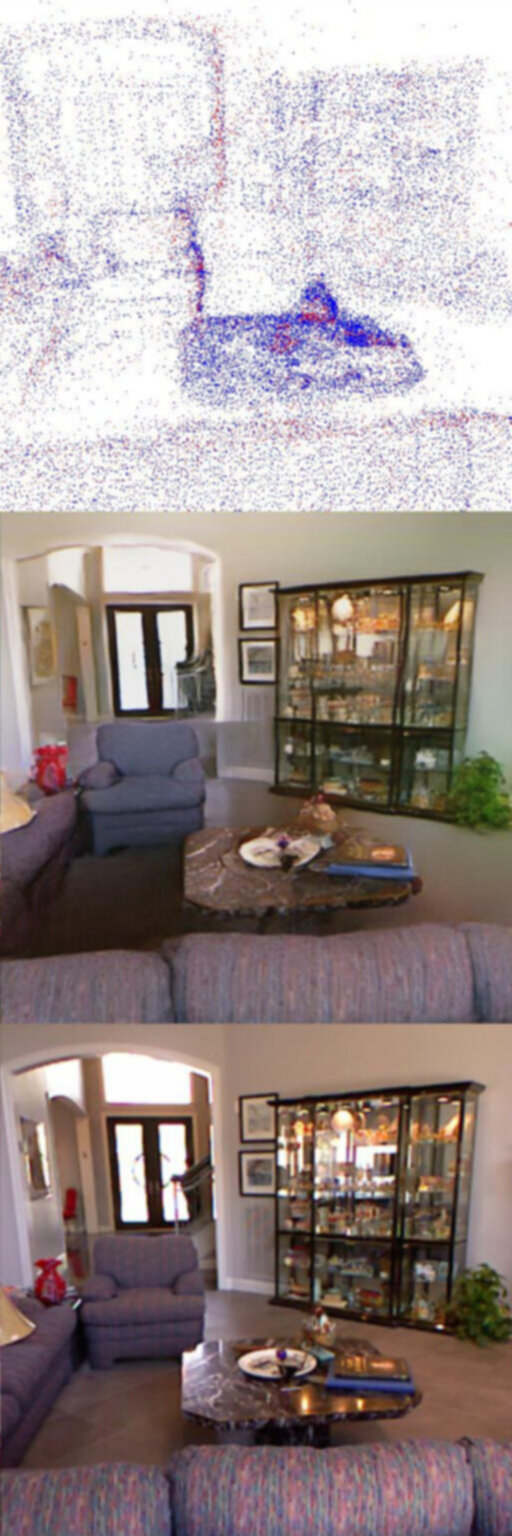}
\includegraphics[width=0.12\linewidth]{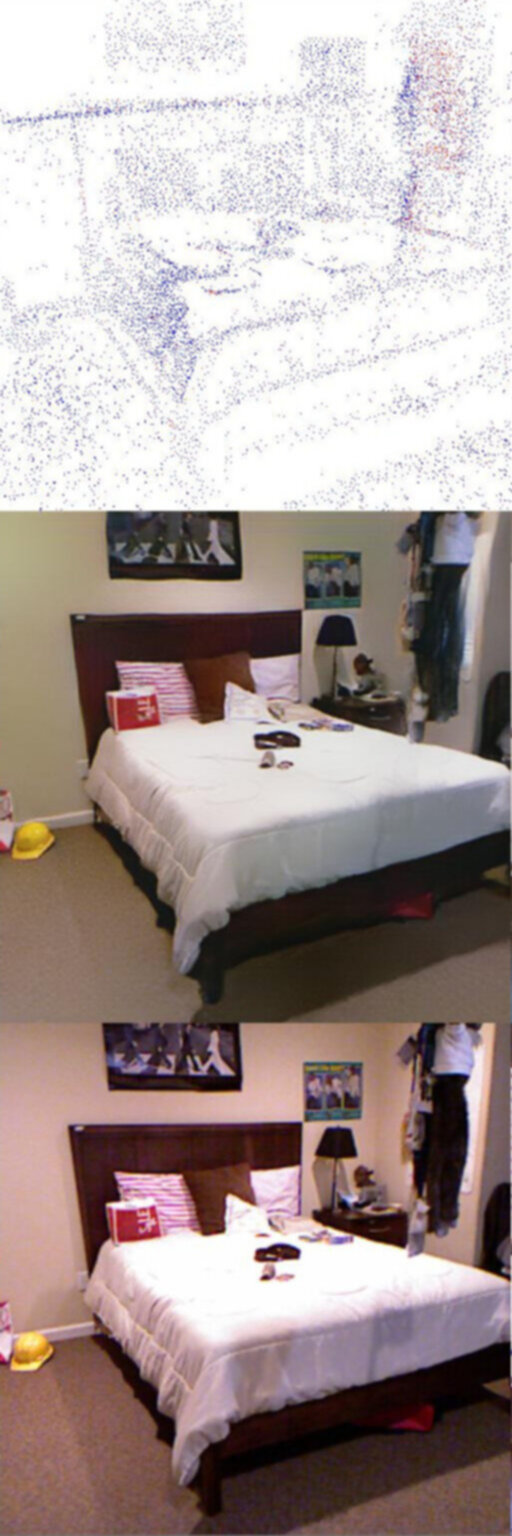}
\includegraphics[width=0.12\linewidth]{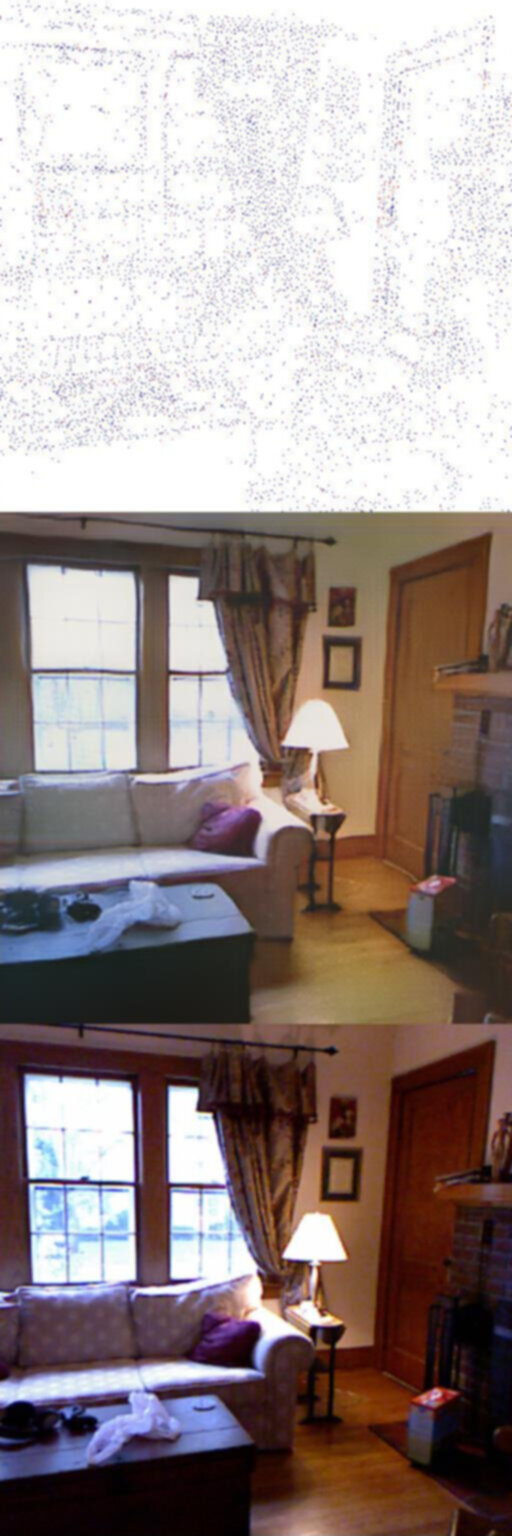}
\includegraphics[width=0.12\linewidth]{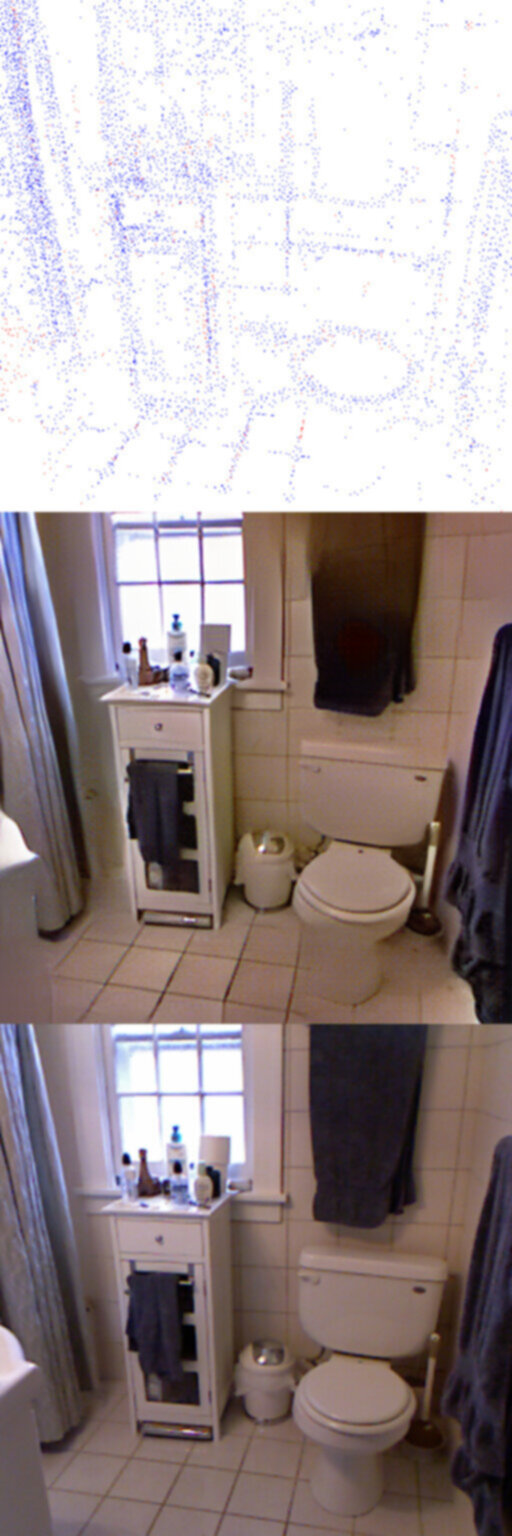}\\
\includegraphics[width=0.12\linewidth]{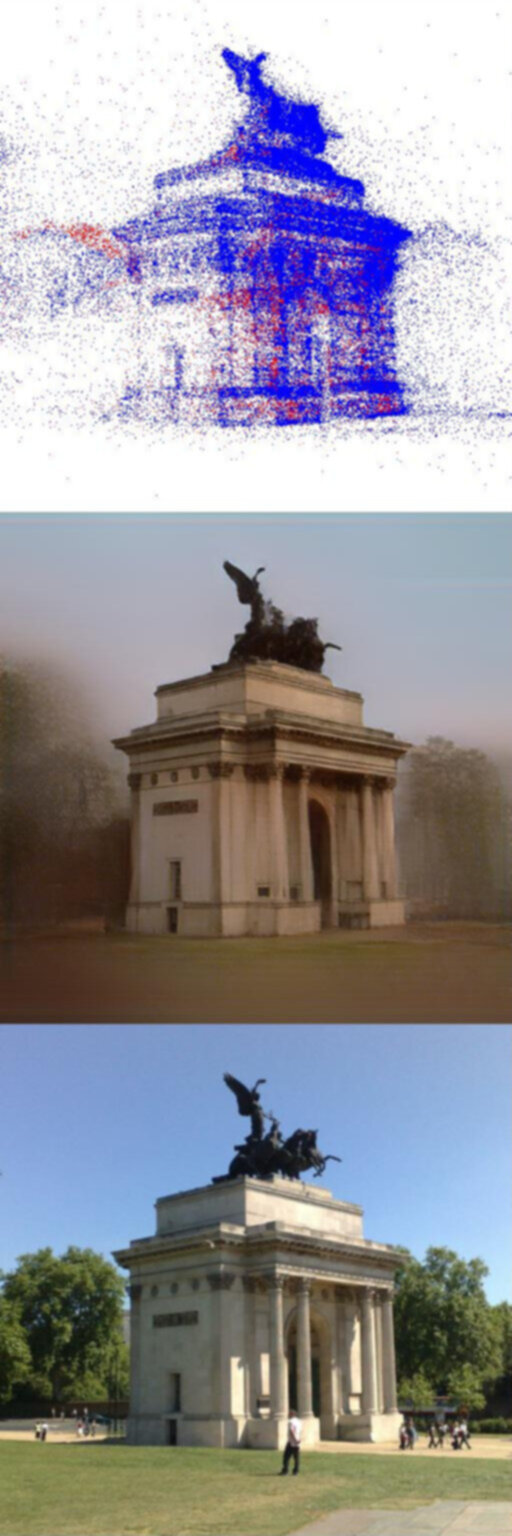}
\includegraphics[width=0.12\linewidth]{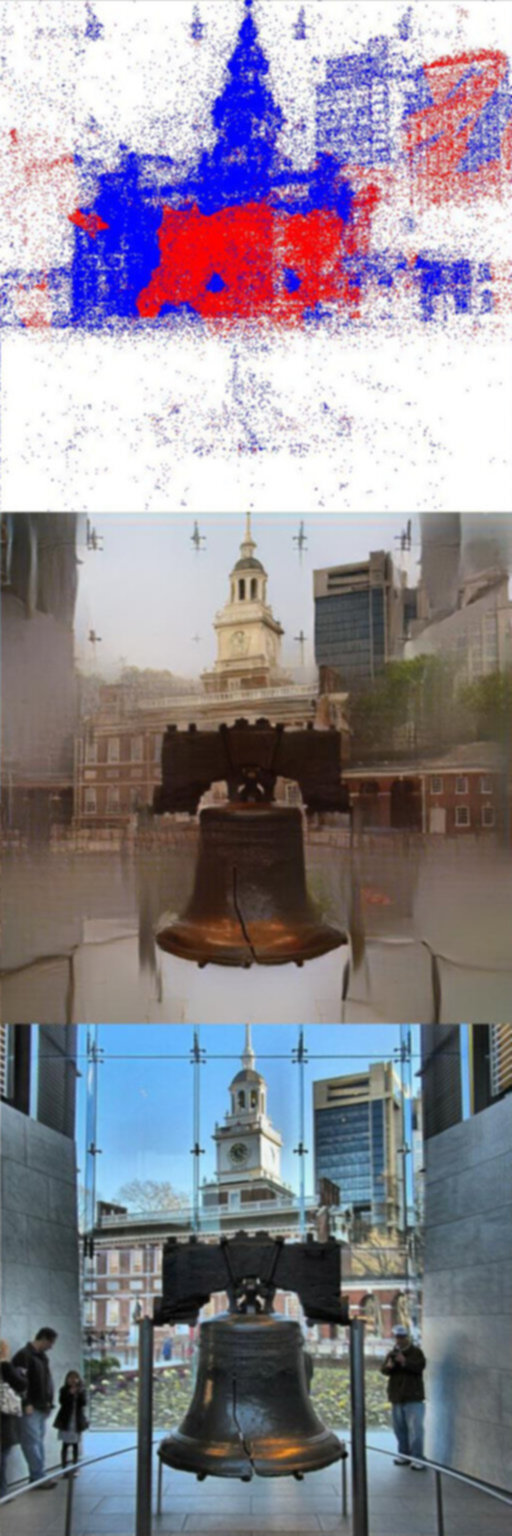}
\includegraphics[width=0.12\linewidth]{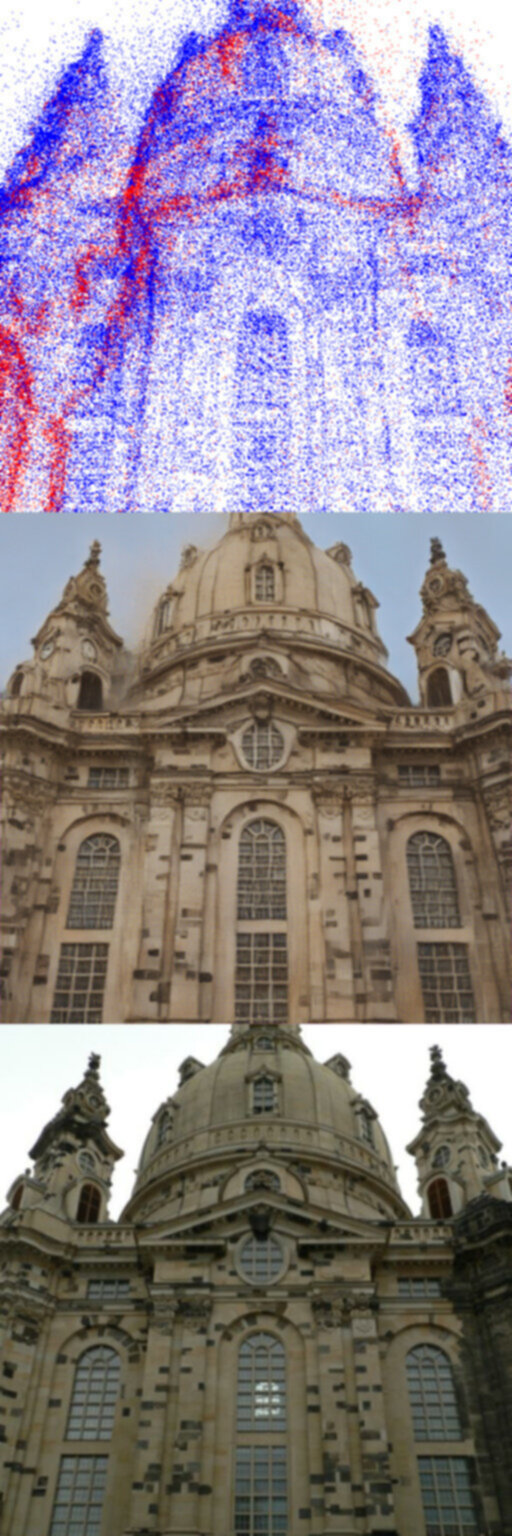}
\includegraphics[width=0.12\linewidth]{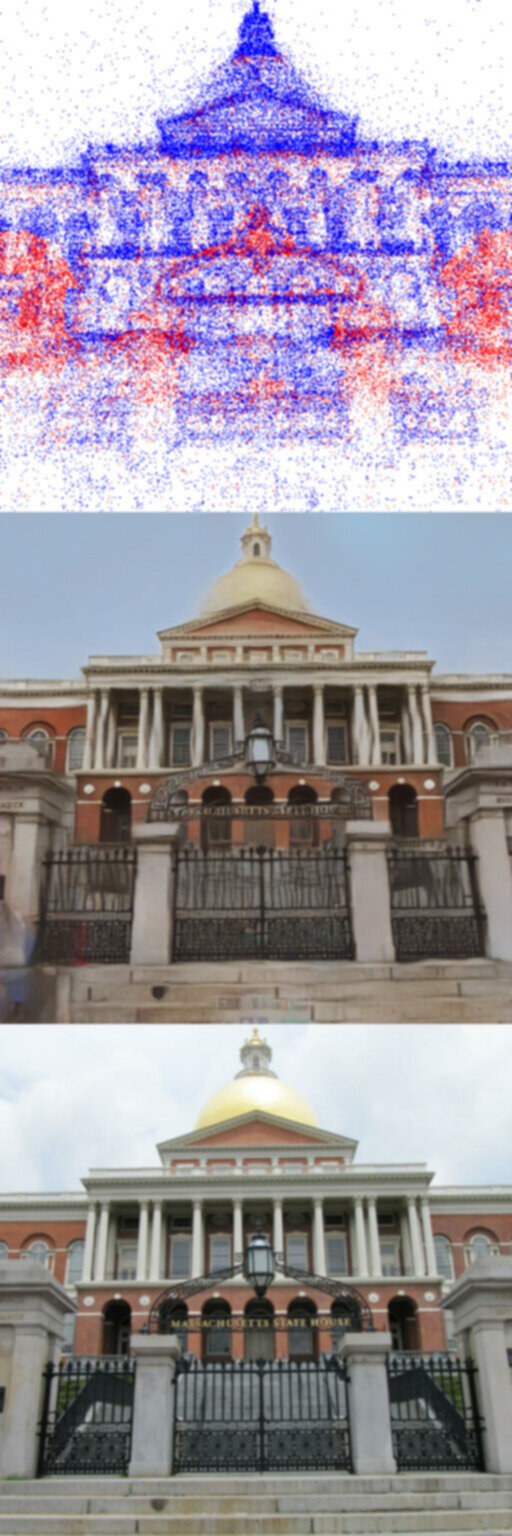}
\includegraphics[width=0.12\linewidth]{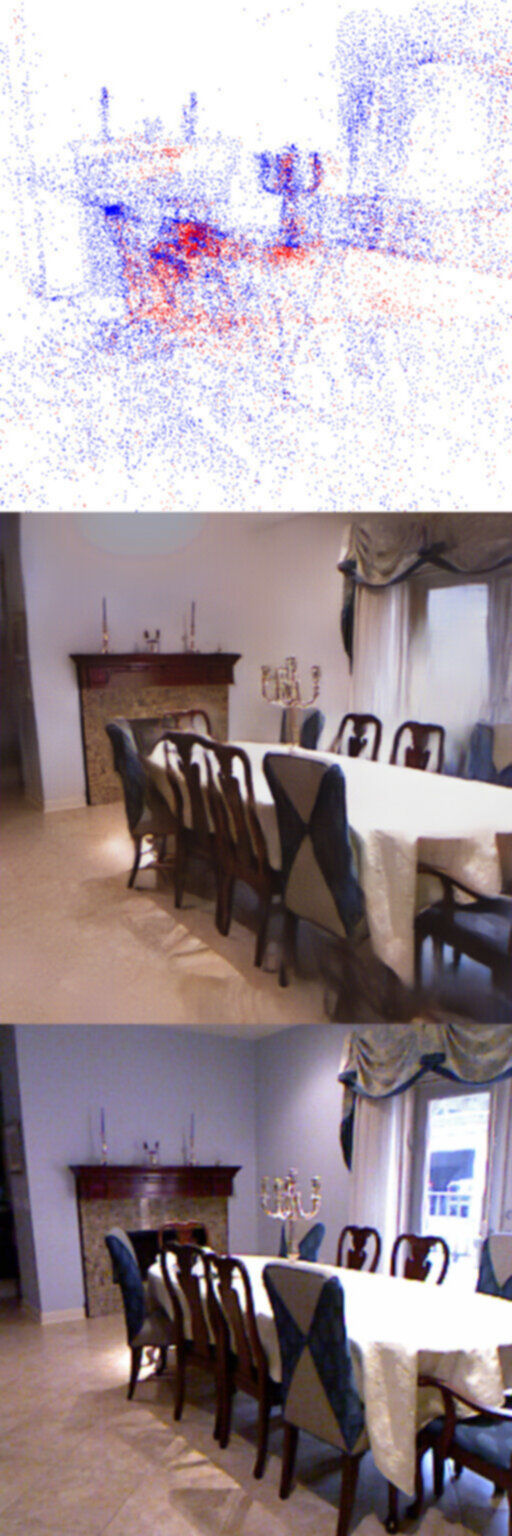}
\includegraphics[width=0.12\linewidth]{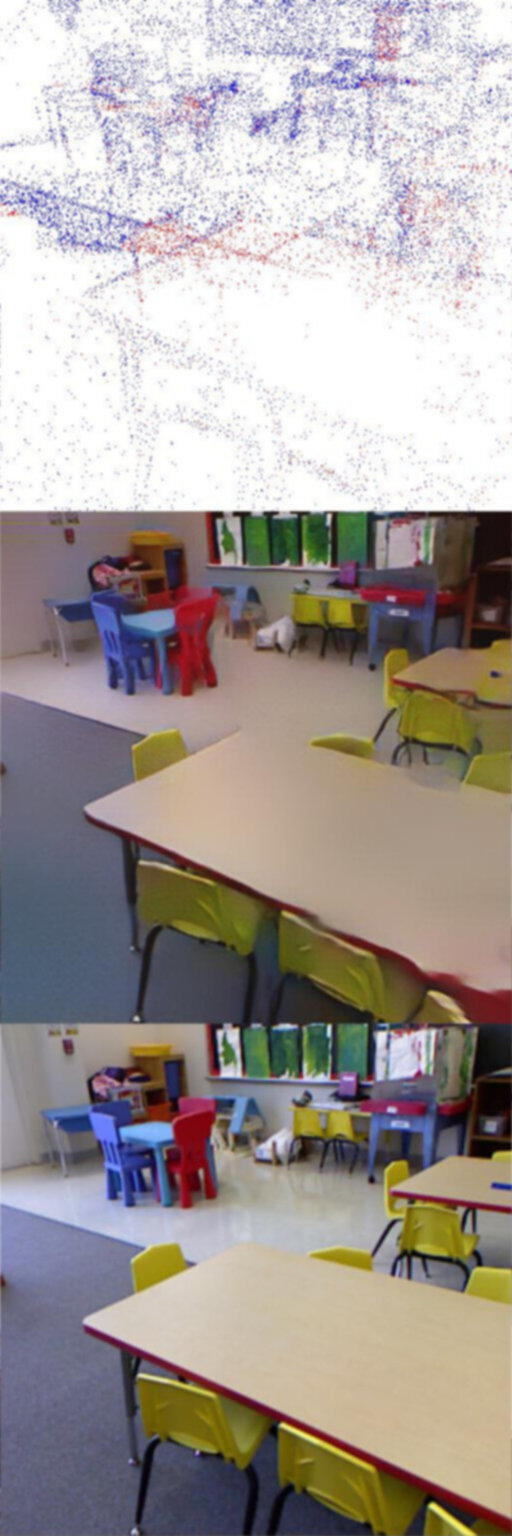}
\includegraphics[width=0.12\linewidth]{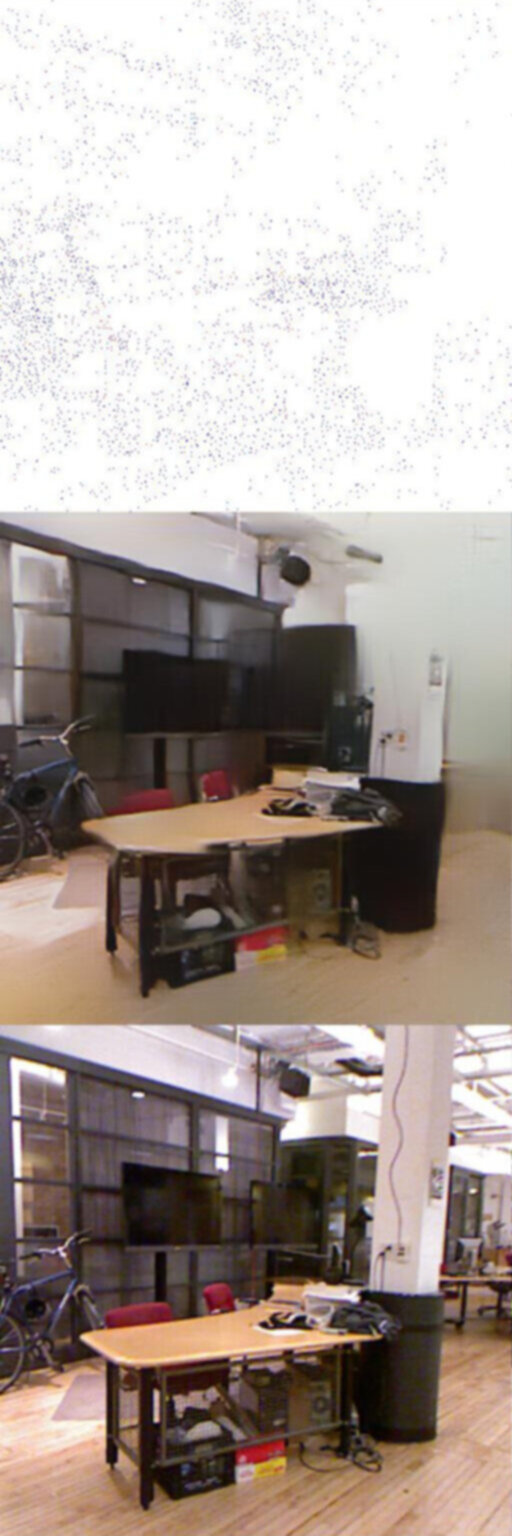}
\includegraphics[width=0.12\linewidth]{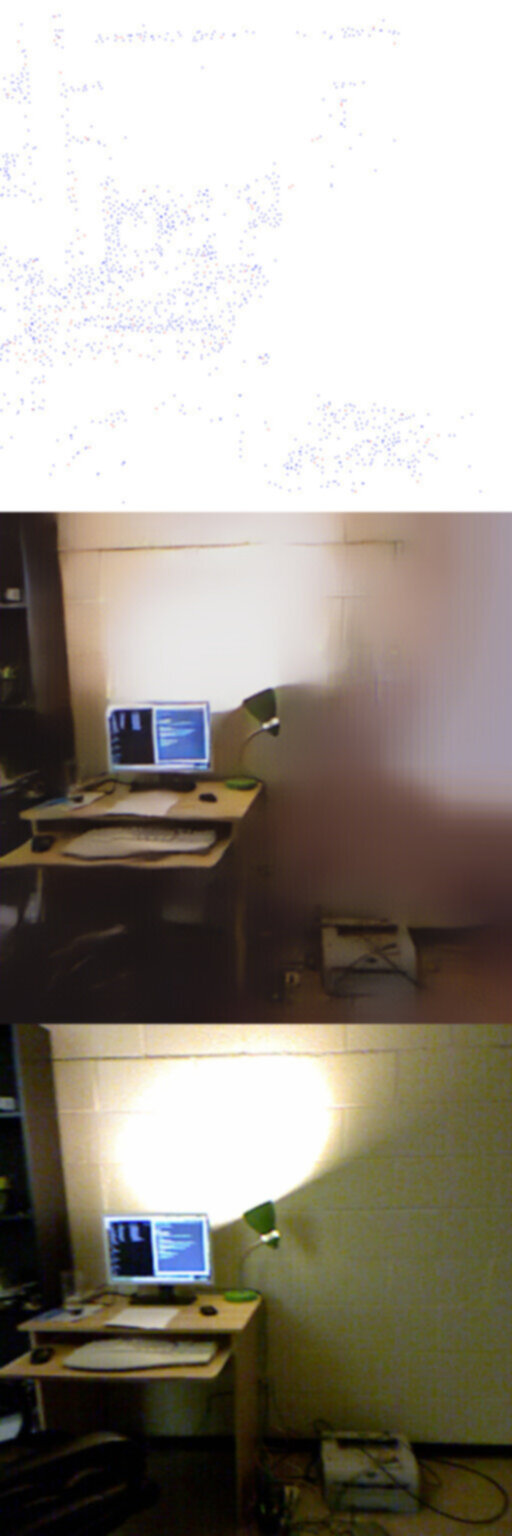}
\caption{\textsc{Qualitative Results:} Each result is a $3 \times 1$ set of square images, showing point clouds (with occluded points in red), image reconstruction and original.  The first four columns (top and bottom) show results from the MegaDepth dataset (internet scenes) and the last four columns (top and bottom) show results from indoor NYU scenes. \textbf{Sparsity:} Our network handles a large variety in input sparsity (density decreases from left to right). In addition, perspective projection accentuates the spatially-varying density differences, and the MegaDepth outdoor scenes have concentrated points in the input whereas NYU indoor scenes have far samples. Further, the input points are non-homogeneous, with large holes which our method gracefully fills in. \textbf{Visual effects:} For the first four columns (MD scenes) our results give the pleasing effect of uniform illumination (see top of first column). Since our method relies on SfM, moving objects are not recovered. \textbf{Scene diversity:} The fourth column is an aerial photograph, an unusual category that is still recovered well. For the last four columns (NYU scenes), despite lower sparsity, we can recover textures in common household scenes such as bathrooms, classrooms and bedrooms. The variety shows that our method does not learn object categories and works on any scene. \textbf{Visibility:} All scenes benefit from
visibility prediction using \textsc{VisibNet} which for example was crucial for the bell example (lower $2^{nd}$ column).}
\vspace{-4mm}
\label{fig_md_nyu}
\end{figure*}
\begin{figure*}
\centering
 \includegraphics[width=0.16\linewidth]{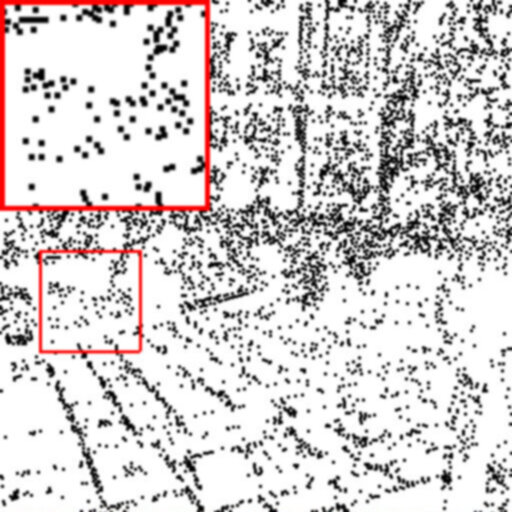}
 \includegraphics[width=0.16\linewidth]{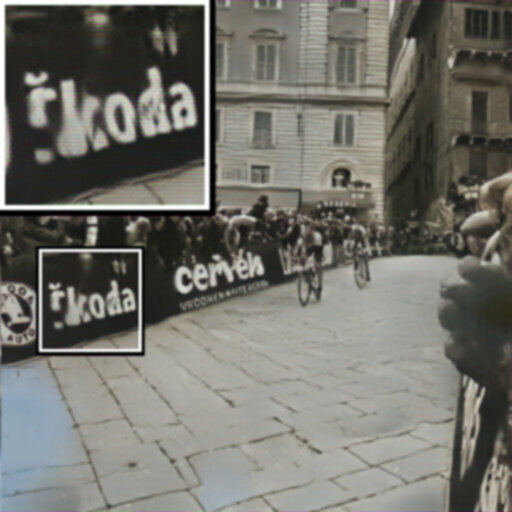}
 \includegraphics[width=0.16\linewidth]{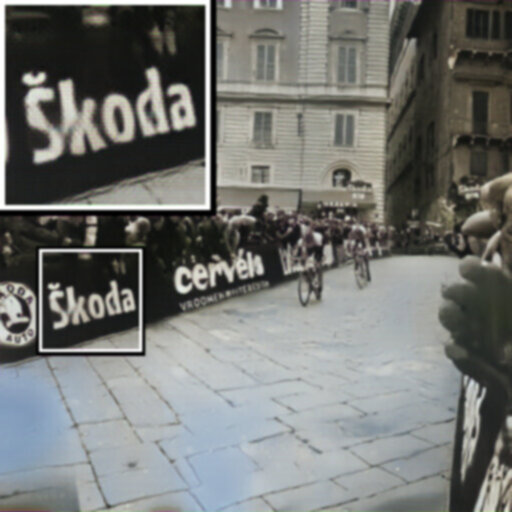}
 \includegraphics[width=0.16\linewidth]{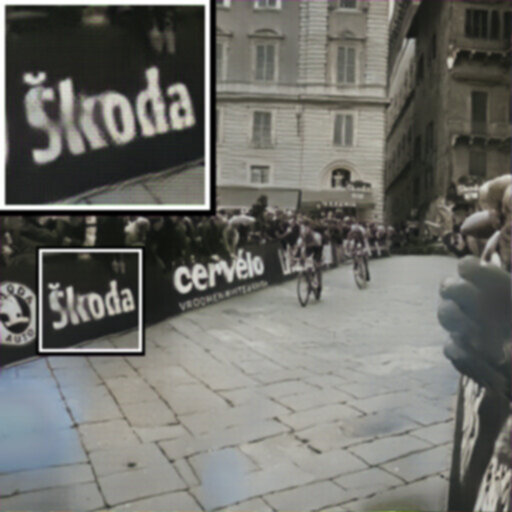}
 \includegraphics[width=0.16\linewidth]{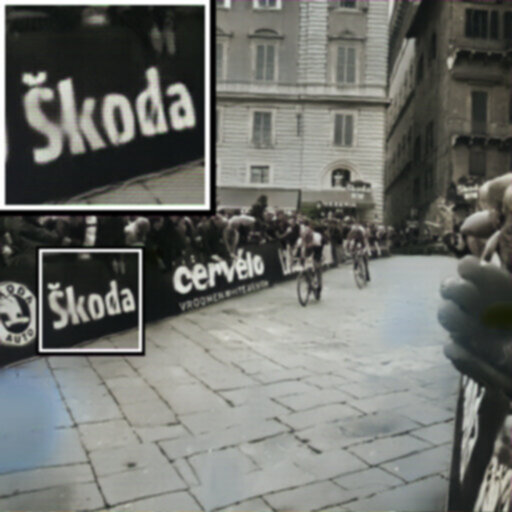}
 \includegraphics[width=0.16\linewidth]{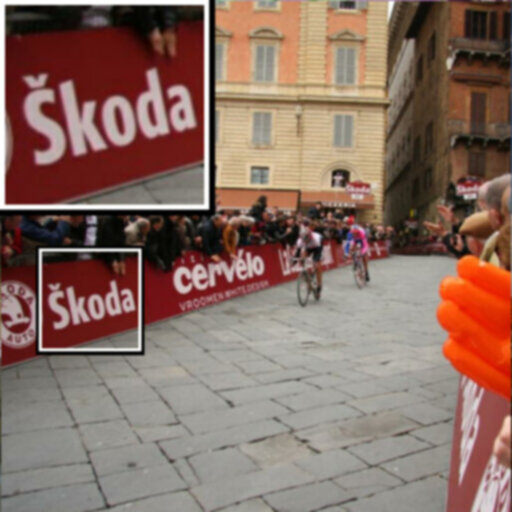}\\
 \vspace{1mm}
\includegraphics[width=0.16\linewidth]{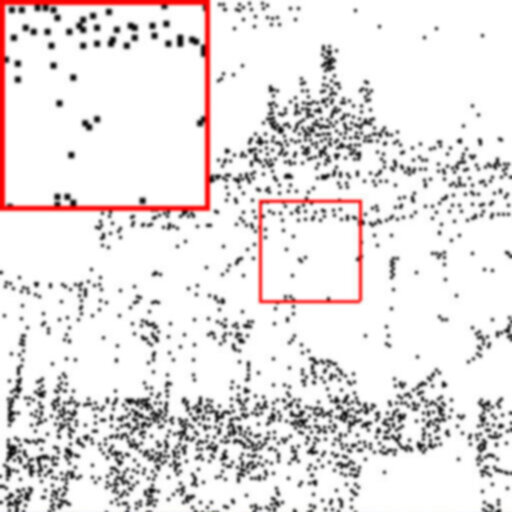}
\includegraphics[width=0.16\linewidth]{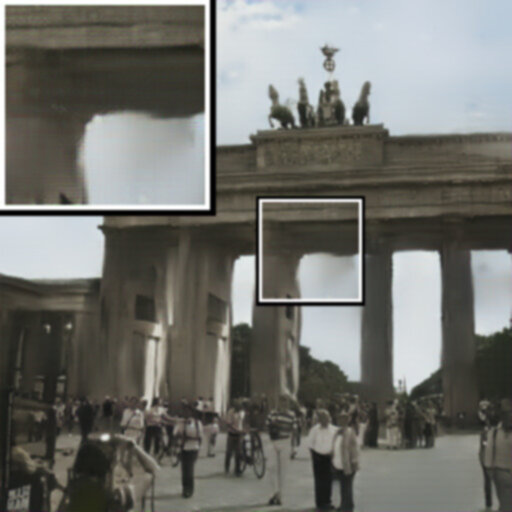}
\includegraphics[width=0.16\linewidth]{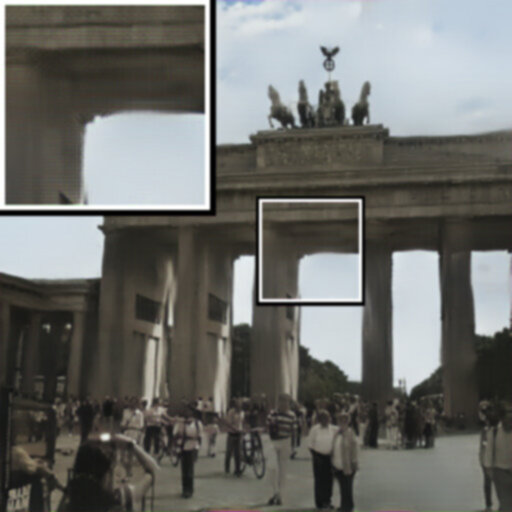}
\includegraphics[width=0.16\linewidth]{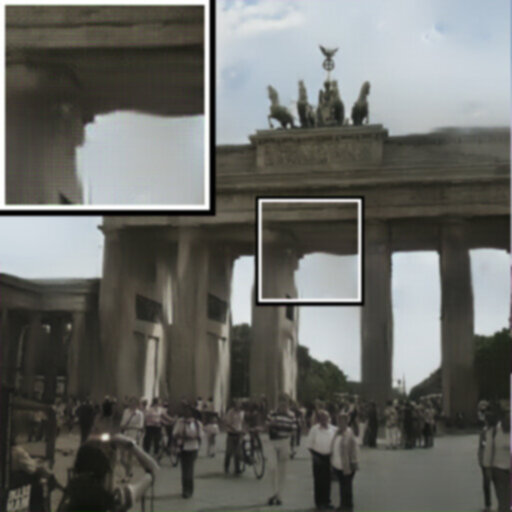}
\includegraphics[width=0.16\linewidth]{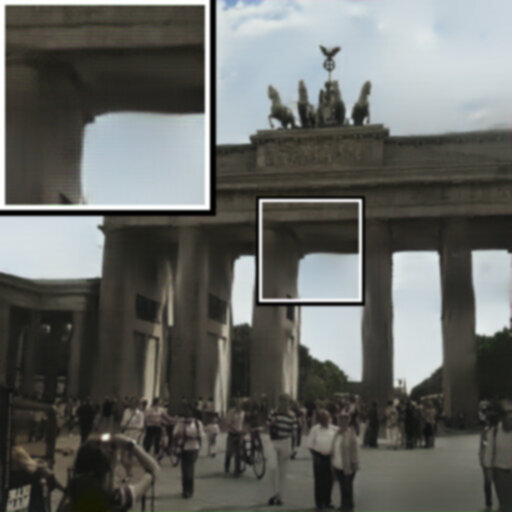}
\includegraphics[width=0.16\linewidth]{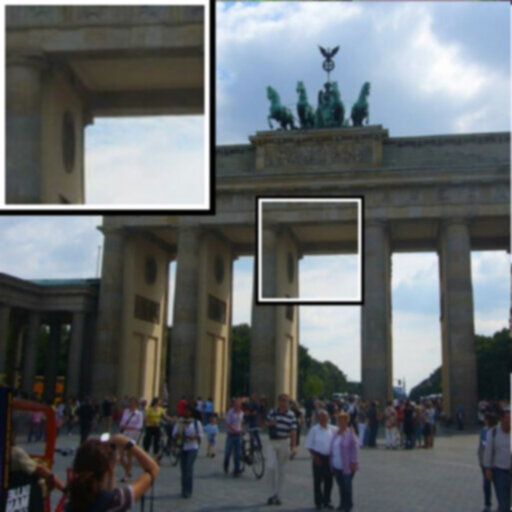}\\
\hspace{1mm} (a) Input \hspace{1.4cm} (b) SIFT \hspace{1.2cm} (c) SIFT + s\hspace{0.8cm} (d) SIFT + o\hspace{0.8cm} (e) SIFT + s + o \hspace{0.8cm} (f) Original \hspace{1mm}
\caption{\textsc{Inverting Sift Features in a Single Image:} (a) 2D keypoint locations. Results obtained with (b) only descriptor, (c) descriptor and keypoint scale, (d) descriptor and keypoint orientation, (e) descriptor, scale and orientation. (f) Original image. Results from using only descriptors ($2^{nd}$ column) are only slightly worse than the baseline ($5^{th}$ column).}
\label{fig:ablation}
\vspace{-1em}
\end{figure*}

\begin{table}
\begin{center}
\centerline{
\resizebox{1\columnwidth}{!}{
\begin{tabular}{c|ccc|ccc|ccc}
\toprule
\textbf{Desc.} & \multicolumn{3}{c|}{\textbf{Inp. Feat.}} & \multicolumn{3}{c|}{\textbf{MAE}} & \multicolumn{3}{c}{\textbf{{}SSIM}}\\
\textbf{Src.} & D & O & S &  20\% & 60\% & 100\% & 20\% & 60\% & 100\% \\
\hline 
\small{Si} & \chk & \chk & \chk & .126 & .105 & .101 & .539 & .605 & .631 \\
\small{Si} & \chk & \chk & \X   & .133 & .111 & .105 & .499 & .568 & .597 \\
\small{Si} & \chk & \X   & \chk & .129 & .107 & .102 & .507 & .574 & .599 \\
\small{Si} & \chk & \X   & \X   & .131 & .113 & .109 & .477 & .550 & .578 \\
\hline
\small{M} & \chk & \X   & \X & .147 & .128 & .123 & .443 & .499 & .524 \\
\bottomrule
\end{tabular}
}
}
\end{center}
\vspace{-5mm}
\caption{\textsc{Inverting Single Image Sift Features:} The top four rows compare networks designed for different subsets of single image (Si) inputs: descriptor (D), keypoint orientation (O) and scale (S). Test error (MAE) and accuracy (SSIM) obtained when 20\%, 60\% and all the SIFT features are used. Lower MAE and higher SSIM values are better. The last row is for when the descriptors originate from multiple (M) different and unknown source images.}
\label{table:ablation}
\end{table}

\section{Experimental Results}

We now report a systematic evaluation of our method. Some of our results are qualitatively summarized in Fig. \ref{fig_md_nyu}, demonstrating robustness to various challenges, namely, missing information in the point clouds, effectiveness of our visibility estimation, and the sparse and irregular distribution of input samples over a large variety of scenes.

\customparagraph{Dataset.} We use the MegaDepth~\cite{MegaDepth} and NYU~\cite{NYU} datasets in our experiments. MegaDepth (MD) is an Internet image dataset with \apprx150k images of 196 landmark scenes obtained from Flickr. NYU contains \apprx400k images of 464 indoor scenes captured with the Kinect (we only used the RGB images). These datasets cover very different scene content, image resolution, and generate very different distribution of SfM points and camera poses.
Generally, NYU scenes produce far fewer SfM points than the MD scenes.

\customparagraph{Preprocessing.} We processed the 660 scenes in MD and NYU using the SfM implementation in COLMAP~\cite{schoenberger2016sfm}. We partitioned the scenes into training, validation, and testing sets
with 441, 80, and 139 scenes respectively. All images of one scene were included only in one of the three groups. We report results using both the average mean absolute error (MAE), where color values are scaled to the range [0,1]. and average structured similarity (SSIM). Note that lower MAE and higher SSIM values indicate better results.

\customparagraph{Inverting Single Image SIFT Features.} Consider the single image scenario, with trivial visibility estimation and identical input to~\cite{dosovitskiy2016inverting}. We performed an ablation study in this scenario, measuring the effect of inverting features with unknown keypoint scale, orientation, and multiple unknown image sources. Four variants of \textsc{CoarseNet} were trained, then tested at three sparsity levels. The results are shown in Table~\ref{table:ablation} and Figure~\ref{fig:ablation}. Table~\ref{table:ablation} reports MAE and SSIM across a combined MD and NYU dataset. The sparsity percentage refers to how many randomly selected features were retained in the input, and our method handles a wide range of sparsity reasonably well. From the examples in Figure~\ref{fig:ablation}, we observe that the networks are surprisingly robust at inverting features with unknown orientation and scale; while the accuracy drops a bit as expected, the reconstructed images are still recognizable. Finally, we quantify the effect of unknown and different image sources for the SIFT features. The last row of Table~\ref{table:ablation} shows that indeed the feature inversion problem becomes harder but the results are still remarkably good. Having demonstrated that our work solves a harder problem than previously tackled, we now report results on inverting SfM points and their features.

\begin{table}
\begin{center}
\centerline{
\resizebox{.7\columnwidth}{!}{
\begin{tabular}{c|ccc|ccc}
\toprule
\multirow{2}{*}{\textbf{Data}} & \multicolumn{3}{c|}{\textbf{Inp. Feat.}} & \multicolumn{3}{c}{\textbf{Accuracy}} \\
& z & D & C & 20\% & 60\% & 100\% \\
\hline
\multirow{4}{*}{MD} &
  \chk & \X   & \X   & .948 & .948 & .946 \\
& \chk & \X   & \chk & .938 & .943 & .941 \\
& \chk & \chk & \X   & .949 & .951 & .948 \\
& \chk & \chk & \chk & .952 & .952 & .950 \\
\hline
\multirow{4}{*}{NYU} &
  \chk & \X   & \X   & .892 & .907 & .908 \\
& \chk & \X   & \chk & .897 & .908 & .910 \\
& \chk & \chk & \X   & .895 & .907 & .909 \\
& \chk & \chk & \chk & .906 & .916 & .917 \\
\bottomrule
\end{tabular}
}
}
\end{center}
\vspace{-2mm}
\caption{\textsc{Evaluation of VisibNet:} We trained four version of \textsc{VisibNet}, each with a different set of input attributes, namely, z (depth), D (SIFT) and C (color) to evaluate their relative importance. Ground truth labels were obtained with VisibDense. The table reports mean classification accuracy on the test set for the NYU and MD datasets. The results show that \textsc{VisibNet} achieves accuracy greater than 93.8\% and  89.2\% on MD and NYU respectively and is not very sensitive to sparsity levels and input attributes.}
\label{table:visibnet}
\vspace{-1em}
\end{table}
\begin{table}
\begin{center}
\centerline{
\resizebox{1\columnwidth}{!}{
\begin{tabular}{c|c|ccc|ccc}
\toprule
\multirow{2}{*}{\textbf{Data}} & \textbf{Visibility} & \multicolumn{3}{c|}{\textbf{MAE}} & \multicolumn{3}{c}{\textbf{SSIM}} \\
& \textbf{Est.} &  20\% & 60\% & 100\% & 20\% & 60\%  & 100\% \\
\hline
\multirow{4}{*}{MD} &
Implicit      & .201 & .197 & .195 & .412 & .436 & .445 \\
& VisibSparse & .202 & .197 & .196 & .408 & .432 & .440 \\
& VisibNet    & .201 & .196 & .195 & .415 & .440 & .448 \\
& VisibDense   & .201 & .196 & .195 & .417 & .442 & .451 \\
\hline
\multirow{4}{*}{NYU} &
Implicit      & .121 & .100 & .094 & .541 & .580 & .592 \\
& VisibSparse & .122 & .100 & .094 & .539 & .579 & .592 \\
& VisibNet    & .120 & .098 & .092 & .543 & .583 & .595 \\
& VisibDense   & .120 & .097 & .090 & .545 & .587 & .600 \\
\bottomrule
\end{tabular}
}
}
\end{center}
\vspace{-2mm}
\caption{\textsc{Importance of Visibility Estimation:} Both sub-tables show results obtained using \textsc{Implicit} \ie
no explicit occlusion reasoning where of burden of visibility estimation implicitly falls on \textsc{CoarseNet},
\emph{VisibNet} and the geometric methods \textsc{VisibSparse} and \textsc{VisibDense}. Lower MAE and higher SSIM values are better.}
\label{table:visibcmp}
\end{table}
\begin{figure*}
\centering
\vspace{5mm}
\includegraphics[width=0.16\linewidth]{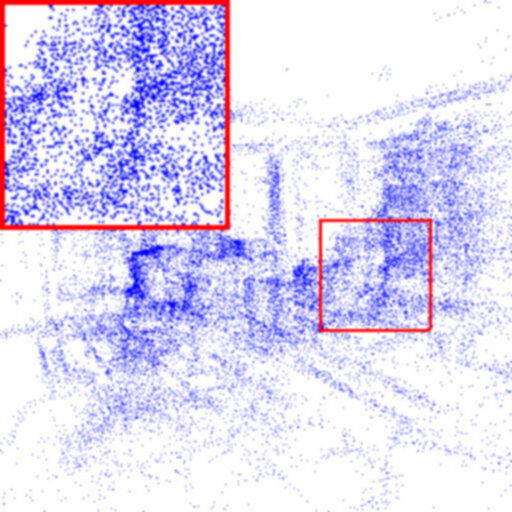}
\includegraphics[width=0.16\linewidth]{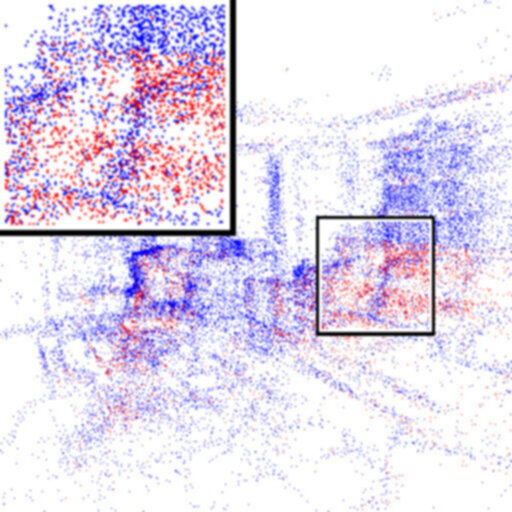}
\includegraphics[width=0.16\linewidth]{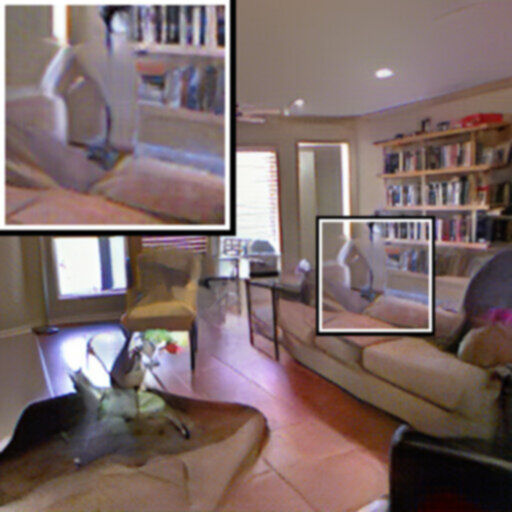}
\includegraphics[width=0.16\linewidth]{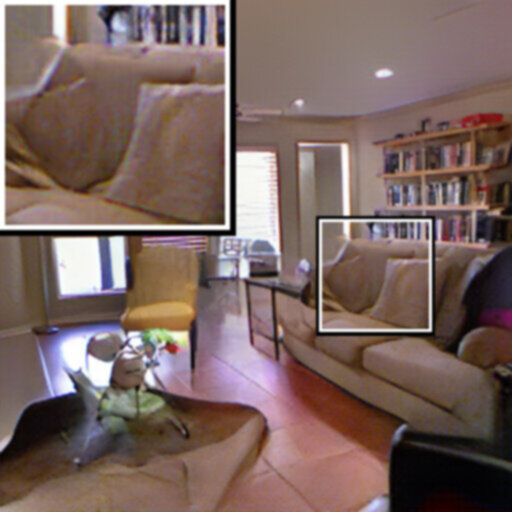}
\includegraphics[width=0.16\linewidth]{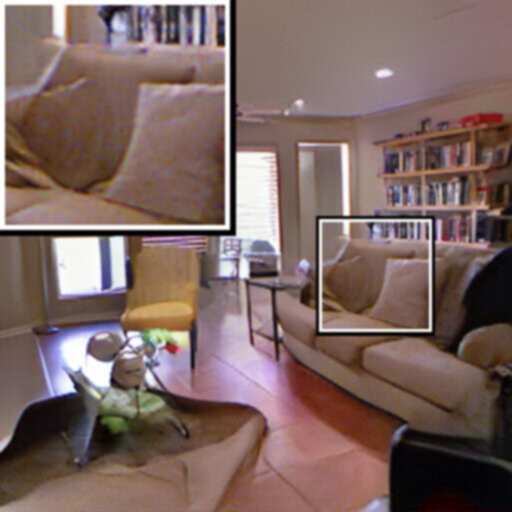}
\includegraphics[width=0.16\linewidth]{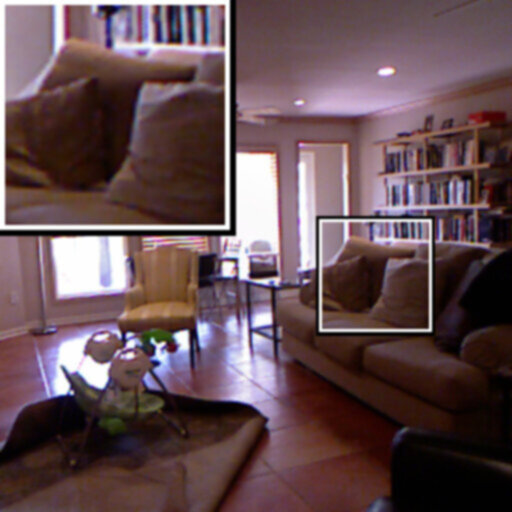}\\
\vspace{1mm}
\includegraphics[width=0.16\linewidth]{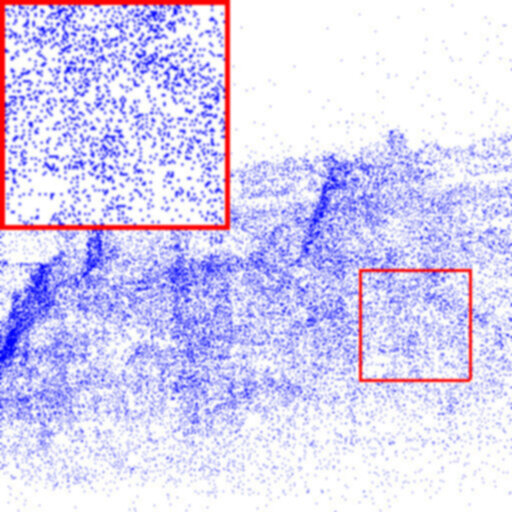}
\includegraphics[width=0.16\linewidth]{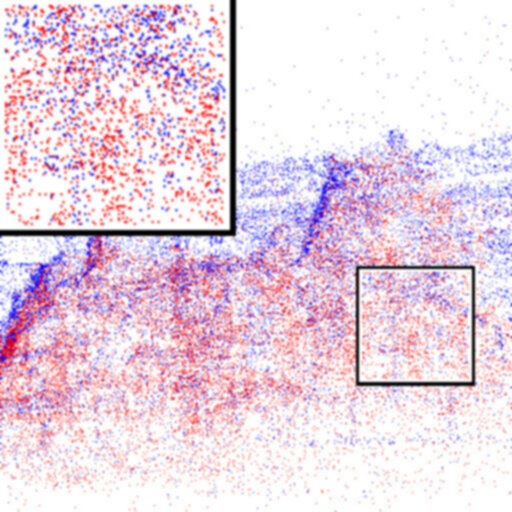}
\includegraphics[width=0.16\linewidth]{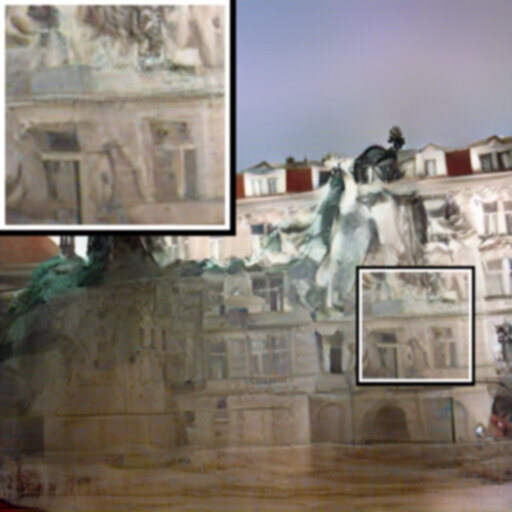}
\includegraphics[width=0.16\linewidth]{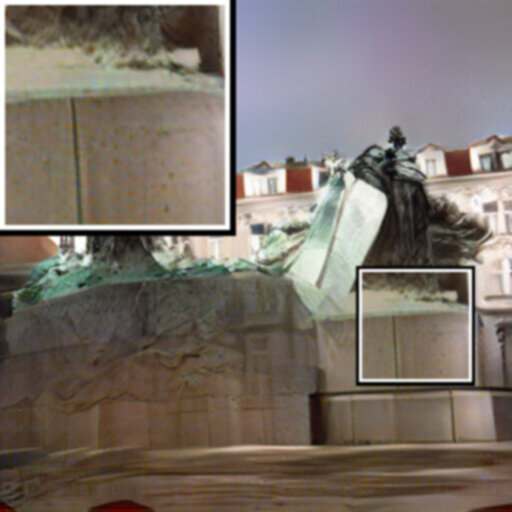}
\includegraphics[width=0.16\linewidth]{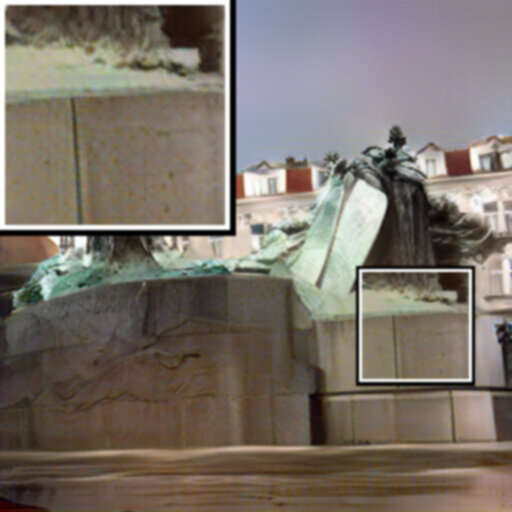}
\includegraphics[width=0.16\linewidth]{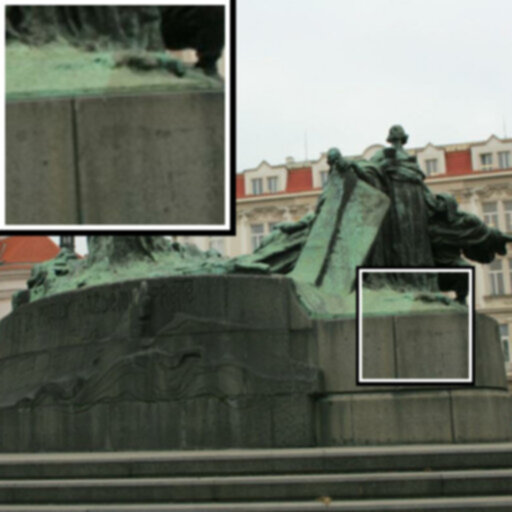}\\
\hspace{3mm} (a) Input \hspace{.7cm} (b) Pred. (VisibNet) \hspace{.5cm} (c) Implicit \hspace{1cm} (d) VisibNet\hspace{1cm} (e) VisibDense \hspace{1cm} (f) Original \hspace{1mm}
\caption{\textsc{Importance of visibility estimation:} Examples showing (a) input 2D point projections (in blue), (b) predicted visibility from \textsc{VisibNet} -- occluded (red) and visible (blue) points, (c--e) results from \textsc{Implicit} (no explicit visibility estimation), \textsc{VisibNet} (uses a CNN) and \textsc{VisibDense} (uses z-buffering and dense models), and (f) the original image.}
\label{fig:visibcmp}
\end{figure*} 

\begin{table}
\begin{center}
\centerline{
\resizebox{1\columnwidth}{!}{
\begin{tabular}{c|ccc|ccc|ccc}
\toprule
\multirow{2}{*}{\textbf{Data}} & \multicolumn{3}{c|}{\textbf{Inp. Feat.}} & \multicolumn{3}{c|}{\textbf{MAE}} & \multicolumn{3}{c}{\textbf{SSIM}}\\
 &  z & D & C &  20\% & 60\% & 100\% & 20\% & 60\% & 100\% \\
\hline % 130k iters
\multirow{4}{*}{MD}
&  \chk & \X   & \X   & .258 & .254 & .253 & .264 & .254 & .250 \\
& \chk & \X   & \chk & .210 & .204 & .202 & .378 & .394 & .403 \\
& \chk & \chk & \X   & .228 & .223 & .221 & .410 & .430 & .438 \\
& \chk & \chk & \chk & .201 & .196 & .195 & .414 & .439 & .448 \\
\hline
\multirow{4}{*}{NYU}
&  \chk & \X   & \X   & .295 & .290 & .289 & .244 & .209 & .197 \\
& \chk  & \X   & \chk & .148 & .121 & .111 & .491 & .528 & .546 \\
& \chk  & \chk & \X   & .207 & .179 & .171 & .493 & .528 & .539 \\
& \chk  & \chk & \chk & .121 & .099 & .093 & .542 & .582 & .594 \\
\bottomrule
\end{tabular}
}
}
\end{center}
\vspace{-2mm}
\caption{\textsc{Effect of Point Attributes:} Performance of four networks designed for different sets of input attributes -- z (depth), D (SIFT) and C (color), on MD and NYU. Input sparsity
is simulated by applying random dropout to input samples during training and testing.} %Here, lower MAE and higher SSIM values are better.}
\label{table:sfm}
\end{table}

\begin{figure}
\centering
\includegraphics[width=0.19\linewidth]{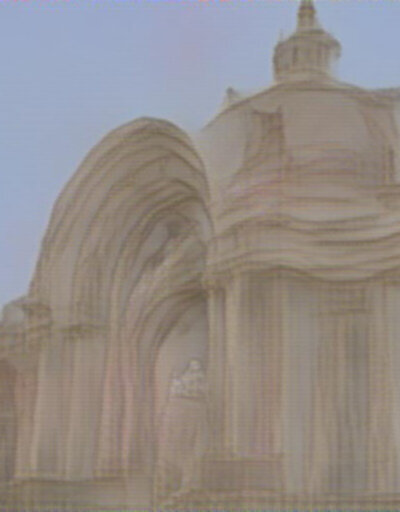}
\includegraphics[width=0.19\linewidth]{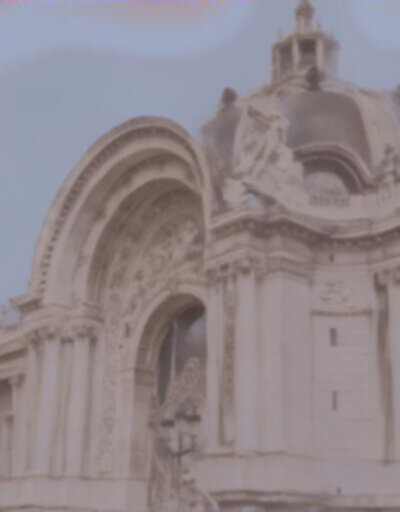}
\includegraphics[width=0.19\linewidth]{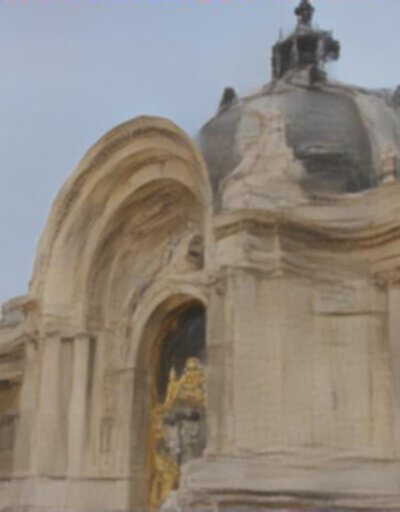}
\includegraphics[width=0.19\linewidth]{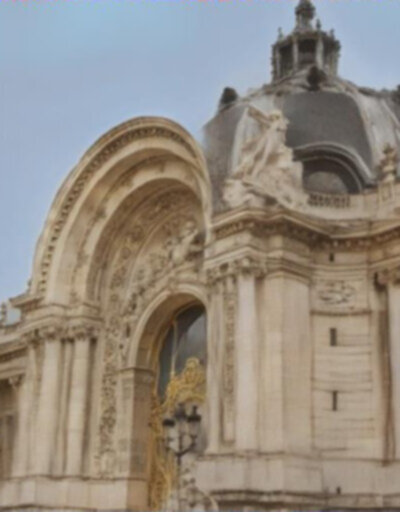}
\includegraphics[width=0.19\linewidth]{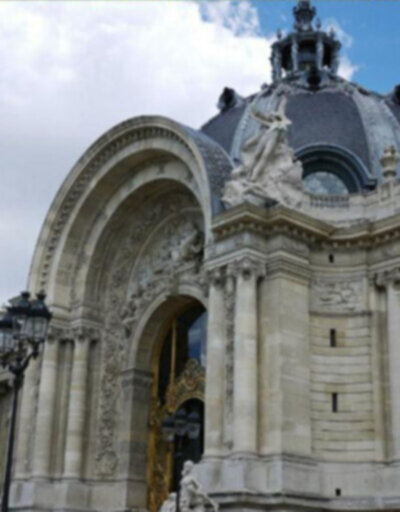}\\
\includegraphics[width=0.19\linewidth]{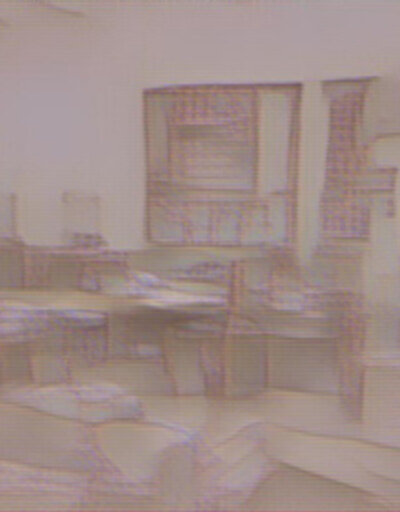}
\includegraphics[width=0.19\linewidth]{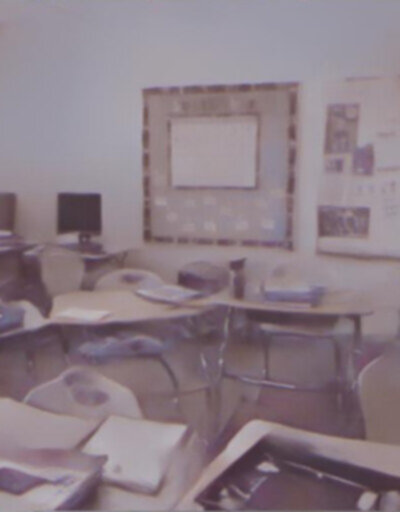}
\includegraphics[width=0.19\linewidth]{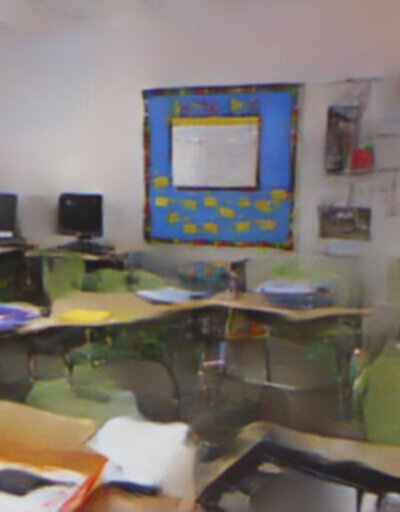}
\includegraphics[width=0.19\linewidth]{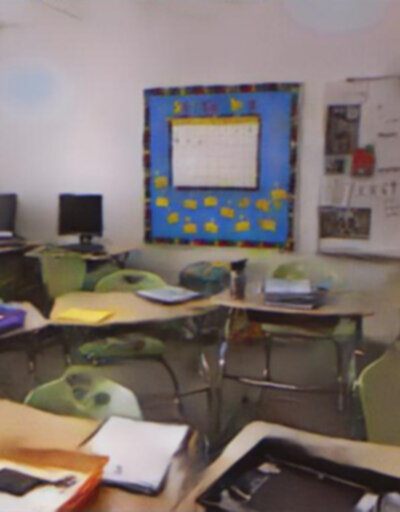}
\includegraphics[width=0.19\linewidth]{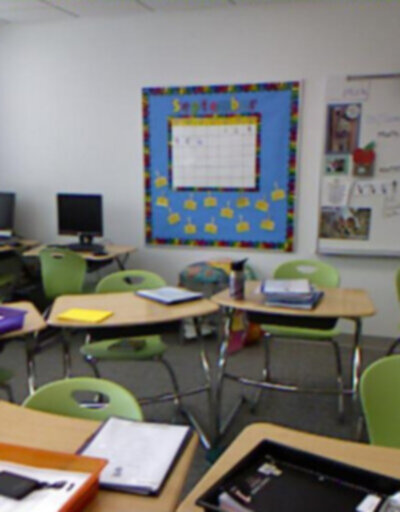}\\
\hspace{2mm} z \hspace{8mm} z + D \hspace{7mm} z + C \hspace{5mm} z + D + C \hspace{5mm} orig\\
\caption{\textsc{Effect of Point Attributes:}
Results obtained with different attributes. Left to right:
depth [z], depth + SIFT [z\,+\,D], depth + color [z\,+\,C], depth + SIFT + color [z\,+\,D\,+\,C] and the original image. (see Table 4).}
\label{fig:input}
\vspace{-1em}
\end{figure}

\begin{figure}
\centering
\includegraphics[width=0.23\linewidth]{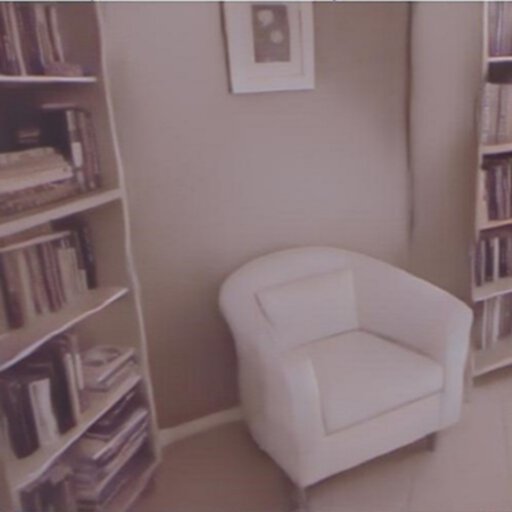}
\includegraphics[width=0.23\linewidth]{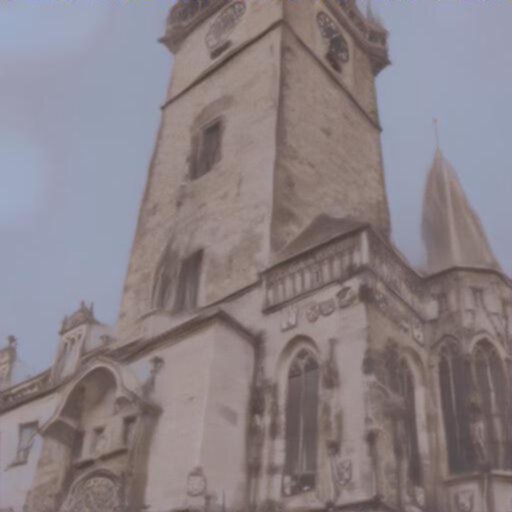}
\hspace{1mm}
\includegraphics[width=0.23\linewidth]{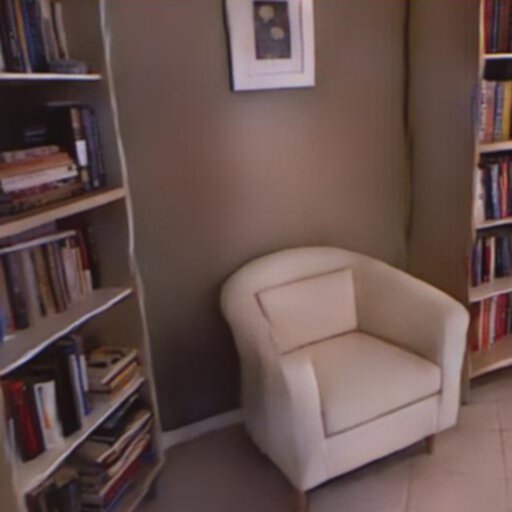}
\includegraphics[width=0.23\linewidth]{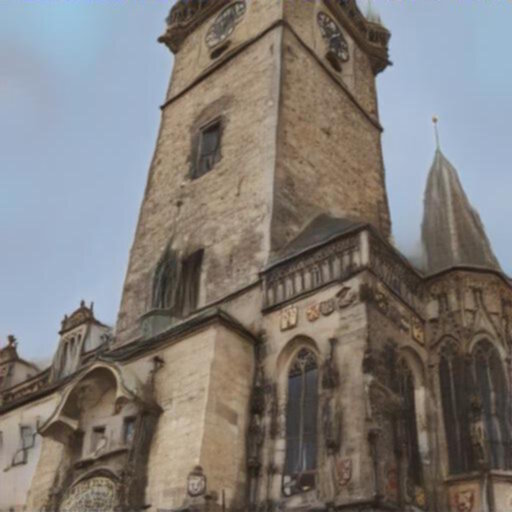}\\
\includegraphics[width=0.23\linewidth]{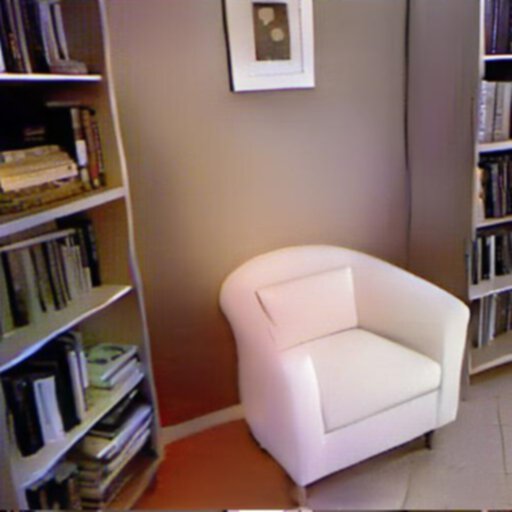}
\includegraphics[width=0.23\linewidth]{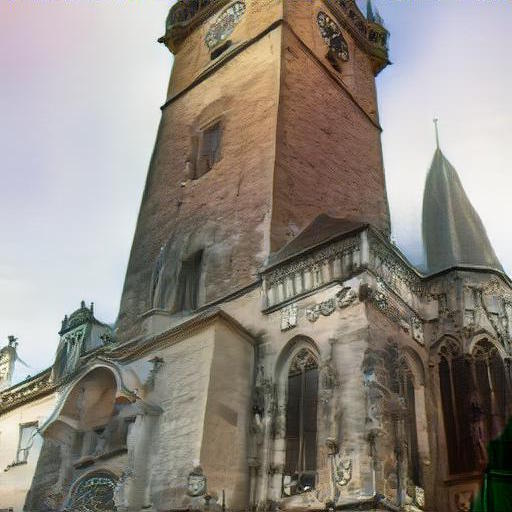}
\hspace{1mm}
\includegraphics[width=0.23\linewidth]{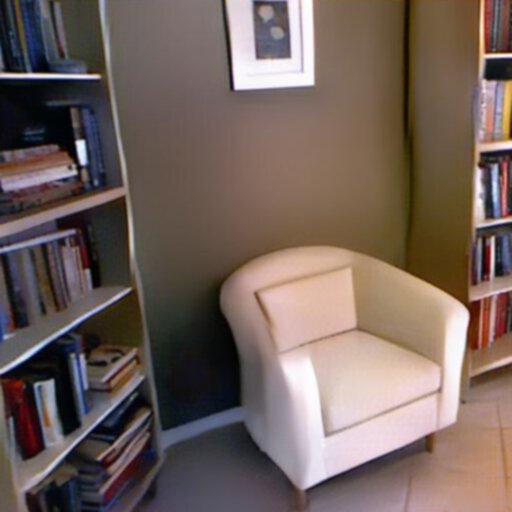}
\includegraphics[width=0.23\linewidth]{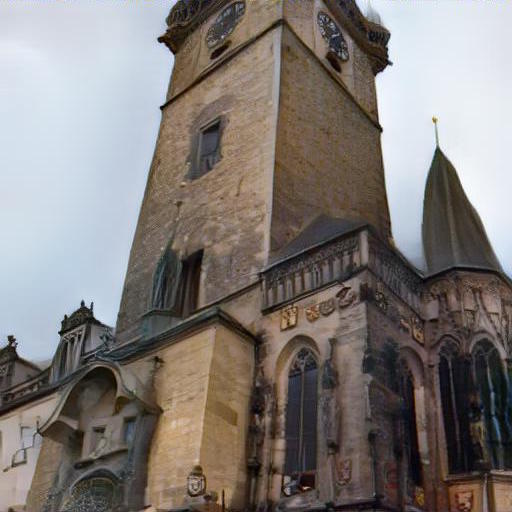}
\hspace{5mm} z + D \hspace{29mm} z + D + C\\
\caption{\textsc{Importance of RefineNet:} (Top row) \textsc{CoarseNet} results. (Bottom Row) \textsc{RefineNet} results.
(Left) Networks use depth and descriptors (z\,+\,D). (Right) Networks use depth, descriptor and color (z\,+\,D\,+\,C).}
\label{fig:refine}
\end{figure} 
\begin{figure}
\centering
\includegraphics[width=\linewidth]{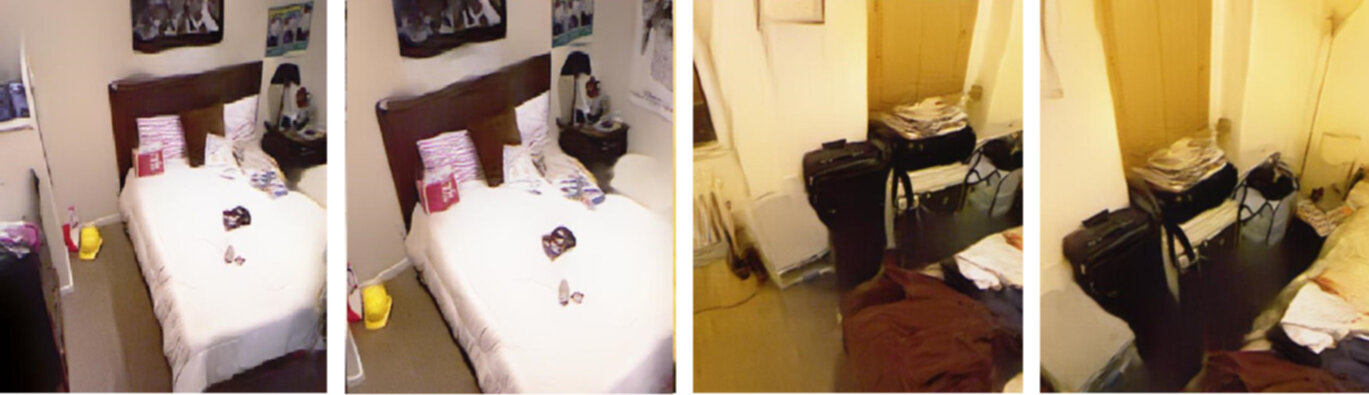}
\caption{\textsc{Novel View Synthesis:} Synthesized images from virtual viewpoints in two NYU scenes~\cite{NYU} helps to interpret the cluttered scenes (see supplementary video).}
\label{fig:novelview}
\vspace{-1em}
\end{figure}

\subsection{Visibility Estimation}

We first independently evaluate the performance of the proposed \textsc{VisibNet} model and compare it to the geometric methods \textsc{VisibSparse} and \textsc{VisibDense}. We trained four variants of \textsc{VisibNet} designed for different subsets of input attributes to
classify points in the input feature map as ``visible'' or ``occluded''. We report classification accuracy separately on the MD and NYU test sets even though the network was trained on the combined training set (see Table~\ref{table:visibnet}). We observe that \textsc{VisibNet} is largely insensitive to scene type, sparsity levels, and choice of input attributes such as depth, color, and descriptors. The \textsc{VisibNet} variant designed for depth only has 94.8\% and 89.2\% mean classification accuracy on MD and NYU test sets, respectively, even when only 20\% of the input samples were used to simulate sparse inputs. Table~\ref{table:visibcmp} shows that when points predicted as occluded by \textsc{VisibNet} are removed from the input to \textsc{CoarseNet}, we observe a consistent improvement when compared to \textsc{CoarseNet} carrying both the burdens of visibility and image synthesis (denoted as \emph{Implicit} in the table). While the improvement may not seem numerically large, in Figure~\ref{fig:visibcmp} we show insets where visual artifacts (bookshelf above, building below) are removed.

\subsection{Relative Significance of Point Attributes}
\vspace{-1mm}

We trained four variants of \textsc{CoarseNet}, each with a different set of the available SfM point attributes. The goal here is to measure the relative importance of each of the attributes. This information could be used to decide which optional attributes should be removed when storing SfM model to enhance privacy. We report reconstruction error on the test set for both indoor (NYU) and outdoor scenes (MD) for various sparsity levels in Table~\ref{table:sfm} and show qualitative evaluation on the test set in Figure~\ref{fig:input}. The results indicate that our approach is largely invariant to sparsity and capable of capturing very fine details even when the input feature map contains just depth, although, not surprisingly, color and SIFT descriptors significantly improves visual quality.

\subsection{Significance of RefineNet}

In Figure~\ref{fig:refine} we qualitatively compare two scenes where the feature maps had only depth and descriptors (left) and when it had all the attributes (right). For privacy preservation, these results are sobering. While Table~\ref{table:sfm} showed that \textsc{CoarseNet} struggles when color is dropped (suggesting an easy solution of removing color for privacy), Figure~\ref{fig:refine} (left) unfortunately shows that \textsc{RefineNet} recovers plausible colors and improves results a lot. Of course, \textsc{RefineNet} trained on all features also does better than \textsc{CoarseNet} although less dramatically (Figure~\ref{fig:refine} (right)).

\subsection{Novel View Synthesis}
Our technique can be used to easily generate realistic novel views of the scene.
While quantitatively evaluating such results is more difficult (in contrast to our experiments where aligned real camera images are available), we show qualitative
results in Figure~\ref{fig:novelview} and generate virtual tours based on the synthesized novel views\footnote{see the video in the supplementary material.}. Such novel view based virtual tours can make scene interpretation easier for an attacker even when the images contain some artifacts.
\section{Conclusion}
In this paper, we introduced a new problem, that of inverting a sparse SfM point cloud and reconstructing color images of the underlying scene. We demonstrated that surprisingly high quality images can be reconstructed from the limited amount of information stored along with sparse 3D point cloud models. Our work highlights the privacy and security risks associated with storing 3D point clouds and the necessity for developing privacy preserving point cloud representations and camera localization techniques, where the persistent scene model data cannot easily be inverted to reveal the appearance of the underlying scene. This was also the primary goal in concurrent work on privacy preserving camera pose estimation~\cite{speciale2019} which proposes a defense against the type of attacks investigated in our paper. Another interesting avenue of future work would be to explore privacy preserving features for recovering correspondences between images and 3D models.

% {\small
% \bibliographystyle{ieee}
% \bibliography{refs.bbl}
% } 

\newpage
\clearpage

\setcounter{page}{1}
\setcounter{section}{0}
\setcounter{figure}{0}
\setcounter{table}{0}
\setcounter{equation}{0}
\setcounter{footnote}{0}
\renewcommand{\thepage}{A\arabic{page}}
\renewcommand{\thesection}{\Alph{section}}
\renewcommand{\thefigure}{A\arabic{figure}}
\renewcommand{\thetable}{A\arabic{table}}
\renewcommand{\theequation}{A\arabic{equation}}

\def\httilde{\mbox{\tt\raisebox{-.5ex}{\symbol{126}}}}

\makeatletter
\def\@thanks{}
\makeatother

\pagenumbering{gobble}

\title{\vspace*{-2.5ex} Supplementary Material: \\ Revealing Scenes by Inverting Structure from Motion Reconstructions \vspace*{-2.0ex}}

\author{Francesco Pittaluga$^{1}$ \quad Sanjeev J.~Koppal$^{1}$ \quad Sing Bing Kang$^{2}$ \quad Sudipta N.~Sinha$^{2}$\\
$^1$ University of Florida \qquad $^2$ Microsoft Research
\vspace{-1.1em}
}

\maketitle

\section{Implementation Details}

In the supplementary material we describe our network architecture and the training procedure in more details.

\subsection{Architecture}

Our network architecture consists of three sub-networks -- \textsc{VisibNet}, \textsc{CoarseNet} and \textsc{RefineNet}. The input to our network is an $H\times W\times n$ dimensional feature map where at each 2D location in the feature map where there is a valid sample, we have one $n$-dimensional feature vector. This feature vector is obtained by concatenating different subsets of depth, color, and SIFT features which are associated with each 3D point in the SfM point cloud. Except for the number of input/output channels in the first/final layers, each sub-network has the same architecture, that of a U-Net with an encoder and a decoder and with skip connections between the layers in the encoder and decoder networks at identical depths. In contrast to conventional U-Nets, where the decoder directly generates the output, in our network, the output of the decoder is passed through three convolutional layers in sequence to obtain the final output.

More, specifically, the architecture of the encoder is {\footnotesize CE256\,-\,CE256\,-\,CE256\,-\,CE512\,-\,CE512\,-\,CE512}, where
CE$N$ denotes a convolutional layer with $N$ kernels of size $4\times4$ and stride equal to 2 followed by an addition of a bias, batch normalization, and a ReLU operation.

The architecture of the decoder is {\footnotesize CD512\,-\,CD512\,-\,CD512\,-\,CD256\,-\,CD256\,-\,CD256\,-\,C128\,-\,C64\,-\,C32-\,C3}, where CD$N$ denotes nearest neighbor upsampling by a factor of 2 followed by a convolutional layer with $N$ kernels of size $3\times3$ and a stride equal to 1, followed by an addition of a bias, batch normalization, and a ReLU operation. C$N$ layers are the same as CD$N$ layers but without the upsampling operation. In the final layer of the decoder, the ReLU is replaced with a tanh non-linearity.
In \textsc{RefineNet}, all ReLU operations are replaced by leaky ReLU operations in all the layers of the decoder network. In \textsc{VisibNet}, the final layer of the decoder has 1 kernel instead of 3.

Our discriminator used for adversarial training of \textsc{RefineNet} has the following architecture --
{\footnotesize CA256\,-\,CA256\,-\,CA256\,-\,CA512\,-\,CA512\,-\,CA512\,-\,FC1024\,-\,FC1024\,-\,FC1024\,-\,FC2}, where
CA$N$ denotes a convolutional layer with $N$ kernels of size $3\times3$ and stride equal to 1 followed by a $2 \times 2$ max pooling operation followed by an addition of a bias, batch normalization, and a leaky ReLU operation. FC$N$ denotes a fully connected layer with $N$ nodes followed by an addition of a bias, and a leaky ReLU operation. In the final layer, the leaky ReLU is replaced by a softmax function.

\subsection{Optimization}

We used the Adam optimizer with $\beta_1=0.9$, $\beta_2=0.999$, $\epsilon = 1\text{e}{-8}$ and a learning rate of 0.0001 for training all networks. Images with resolution $256 \times 256$ pixels were used as input to the network during training. However, the trained network was used to process images at a resolution of $512 \times 512$ pixels. During training, we resized each image such that the smaller dimension of the resized image was randomly assigned to either 296, 394, or 512 pixels, after which we applied a random 2D cropping to the resized image to obtain a $256 \times 256$ image which was the actual input to our network. We used Xavier initialization for all the parameters of our network.

\section{Additional Results}
\begin{figure*}
\centering
\fbox{\includegraphics[width=0.18\linewidth, height=0.18\linewidth]{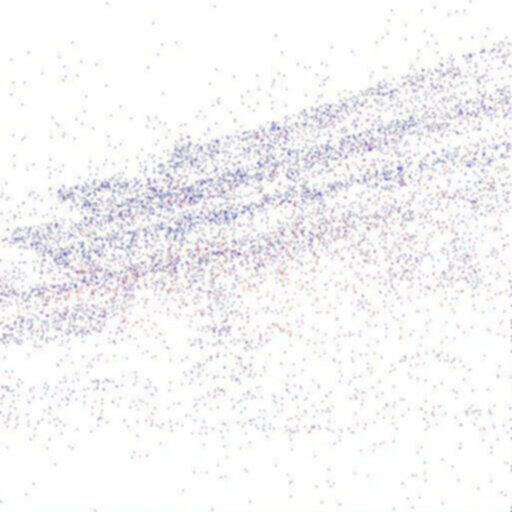}}
\fbox{\includegraphics[width=0.18\linewidth, height=0.18\linewidth]{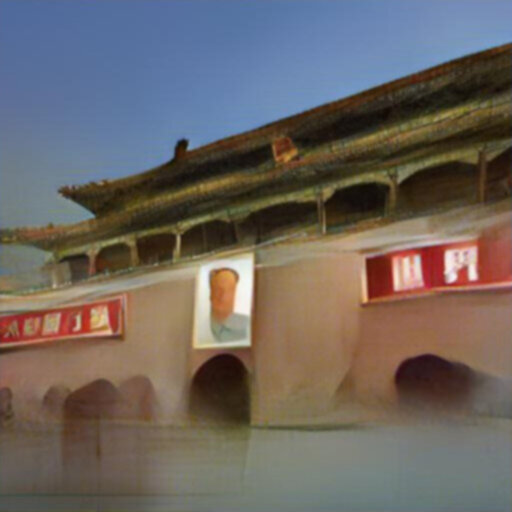}}
\fbox{\includegraphics[width=0.18\linewidth, height=0.18\linewidth]{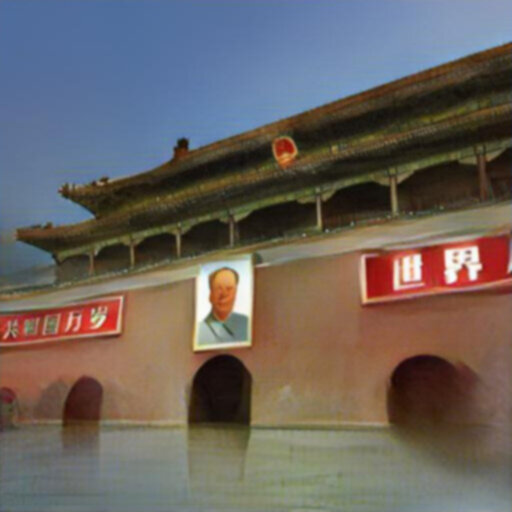}}
\fbox{\includegraphics[width=0.18\linewidth, height=0.18\linewidth]{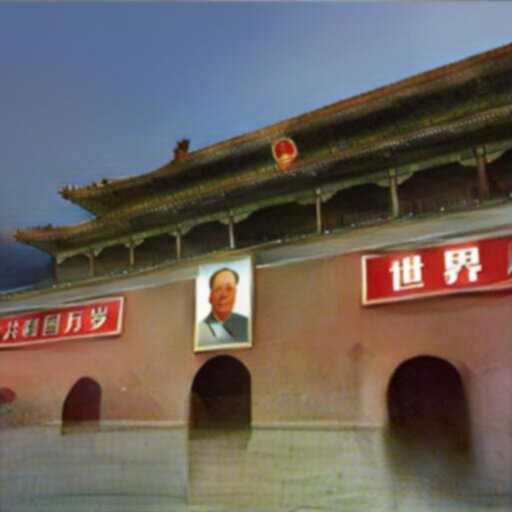}}
\fbox{\includegraphics[width=0.18\linewidth, height=0.18\linewidth]{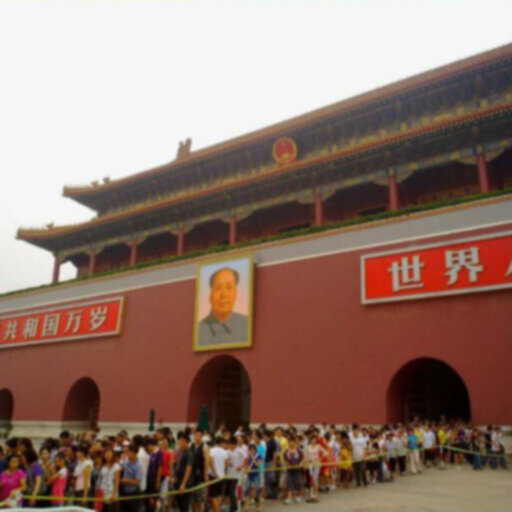}}\\
\vspace{2mm}
\fbox{\includegraphics[width=0.18\linewidth, height=0.18\linewidth]{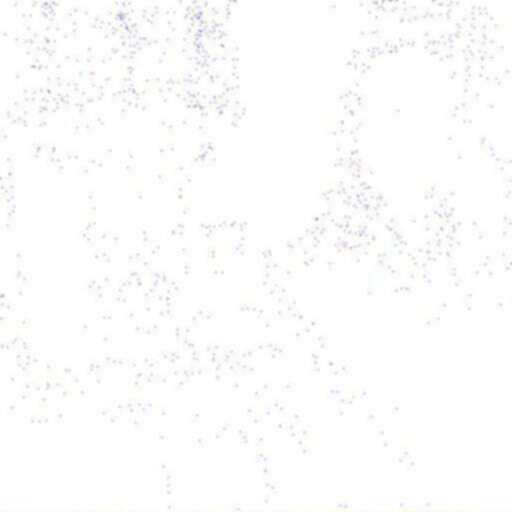}}
\fbox{\includegraphics[width=0.18\linewidth, height=0.18\linewidth]{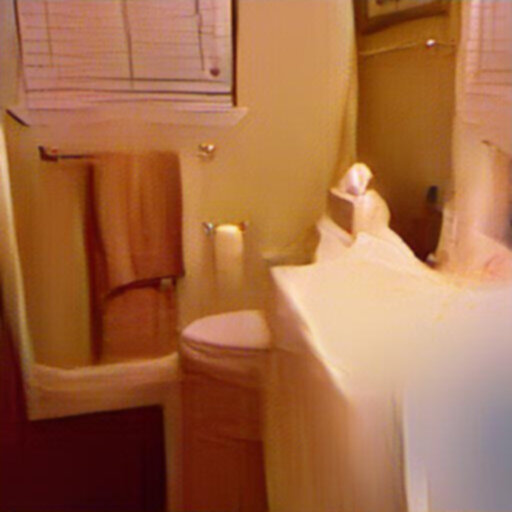}}
\fbox{\includegraphics[width=0.18\linewidth, height=0.18\linewidth]{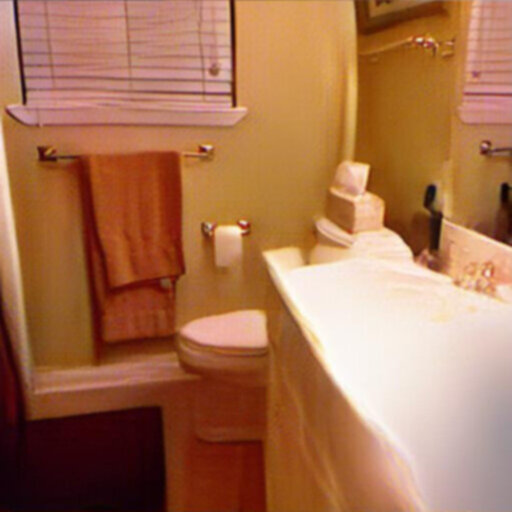}}
\fbox{\includegraphics[width=0.18\linewidth, height=0.18\linewidth]{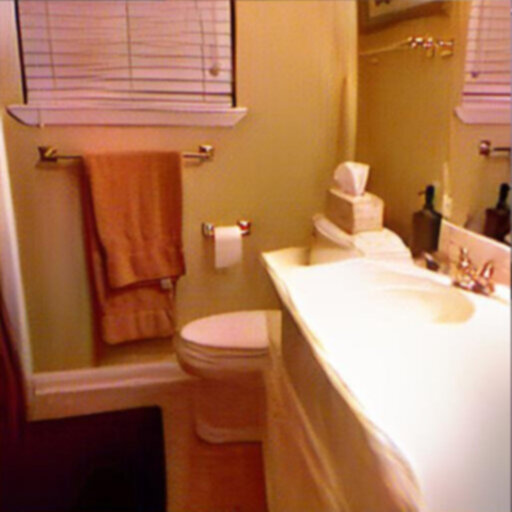}}
\fbox{\includegraphics[width=0.18\linewidth, height=0.18\linewidth]{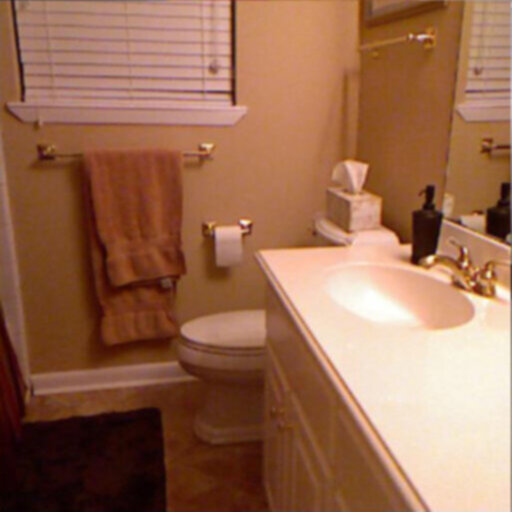}}\\
\vspace{2mm}
(a) Sparse Input (20\%) \hspace{4mm} (b) Using 20\% \hspace{1.2cm} (c) Using 60\% \hspace{1.4cm} (d) Using all \hspace{1.2cm} (e) Original Images
\caption{\textsc{Evaluating Robustness to Sparsity:} Two sets of images synthesized using our complete pipeline, by running \textsc{VisibNet}, \textsc{CoarseNet} and \textsc{RefineNet}. From left to right: (a) Simulated sparse inputs to our networks. Here, only 20\% of the 3D points in the respective SfM models were used. Image synthesized using our method using (b) 20\% of the points, (c) 60\% of the points, (d) all the points and (e) the original source images. Even when the inputs are extremely sparse, most of the contents of the synthesized images can be easily recognized.}
\label{fig:sparsity}
\end{figure*}

\begin{figure*}
\centering
\fbox{\includegraphics[width=0.10\linewidth,height=0.3\linewidth]{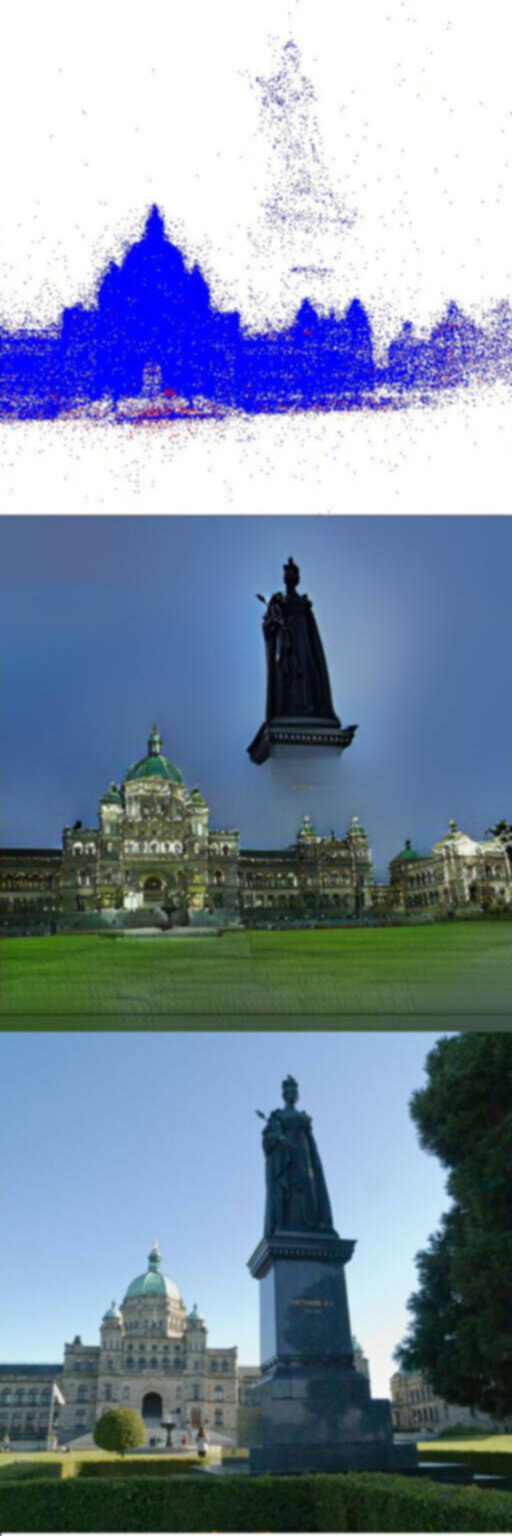}}
\fbox{\includegraphics[width=0.10\linewidth,height=0.3\linewidth]{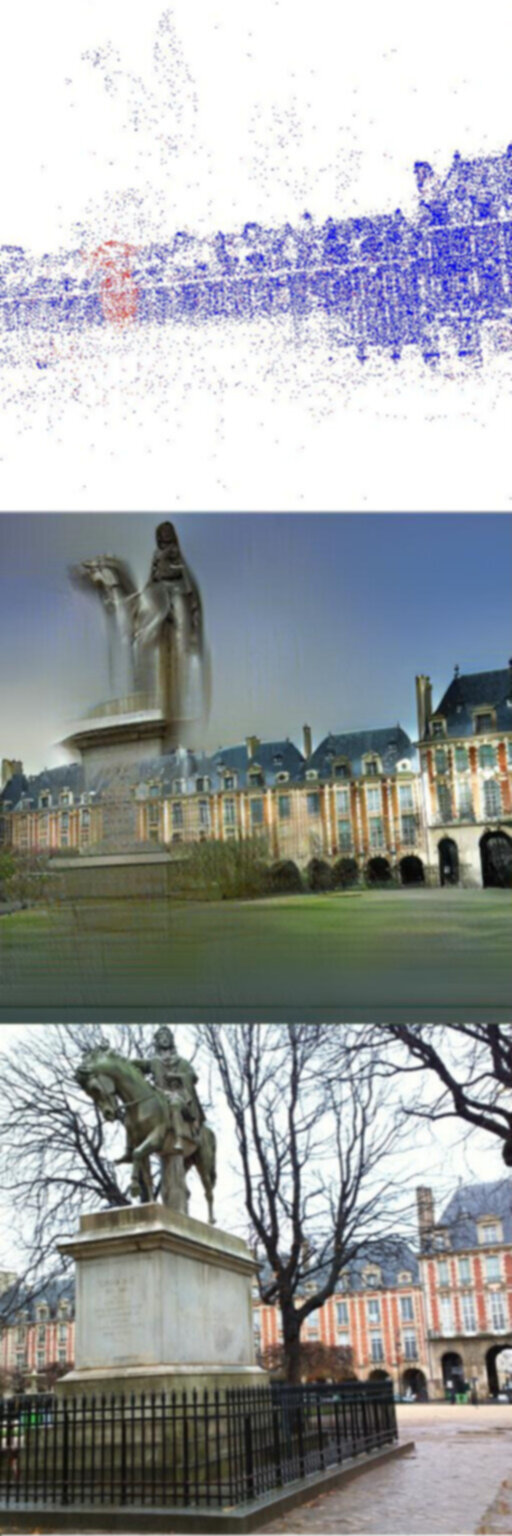}}
\fbox{\includegraphics[width=0.10\linewidth,height=0.3\linewidth]{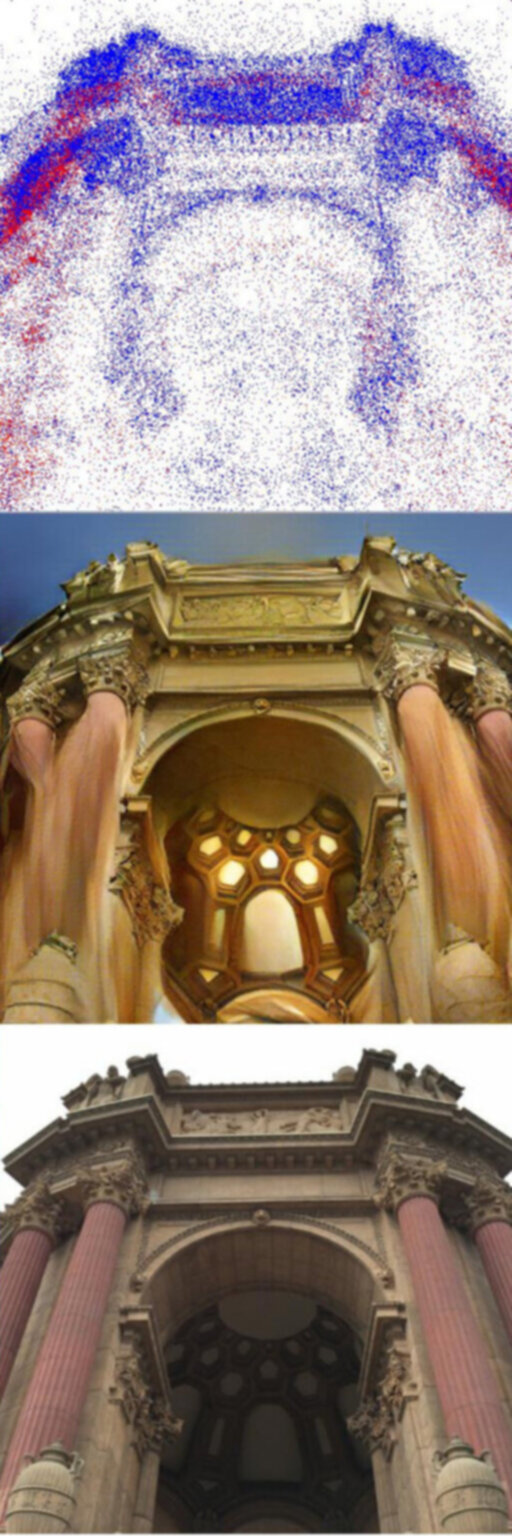}}
\fbox{\includegraphics[width=0.10\linewidth,height=0.3\linewidth]{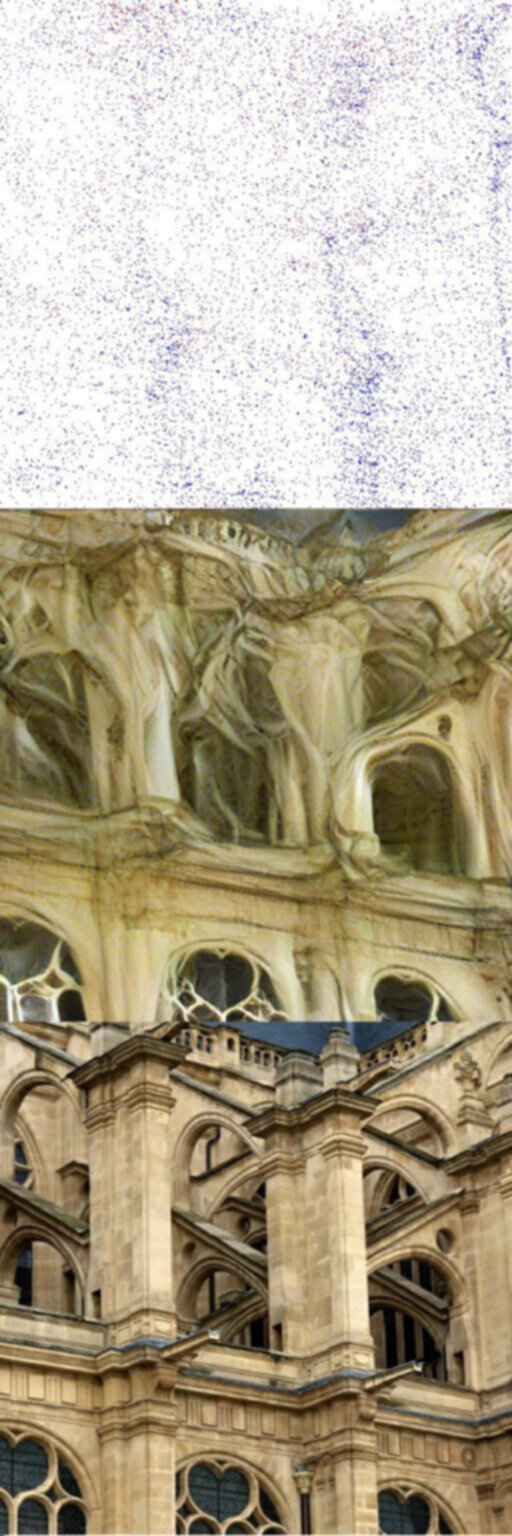}}
\fbox{\includegraphics[width=0.10\linewidth,height=0.3\linewidth]{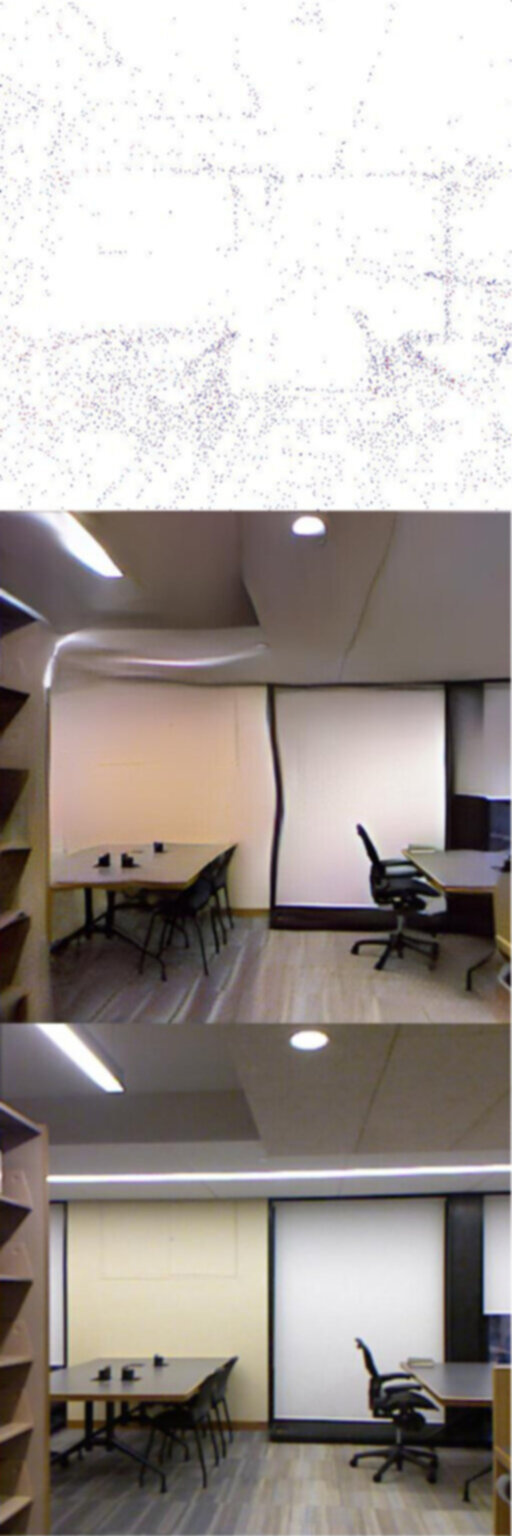}}
\fbox{\includegraphics[width=0.10\linewidth,height=0.3\linewidth]{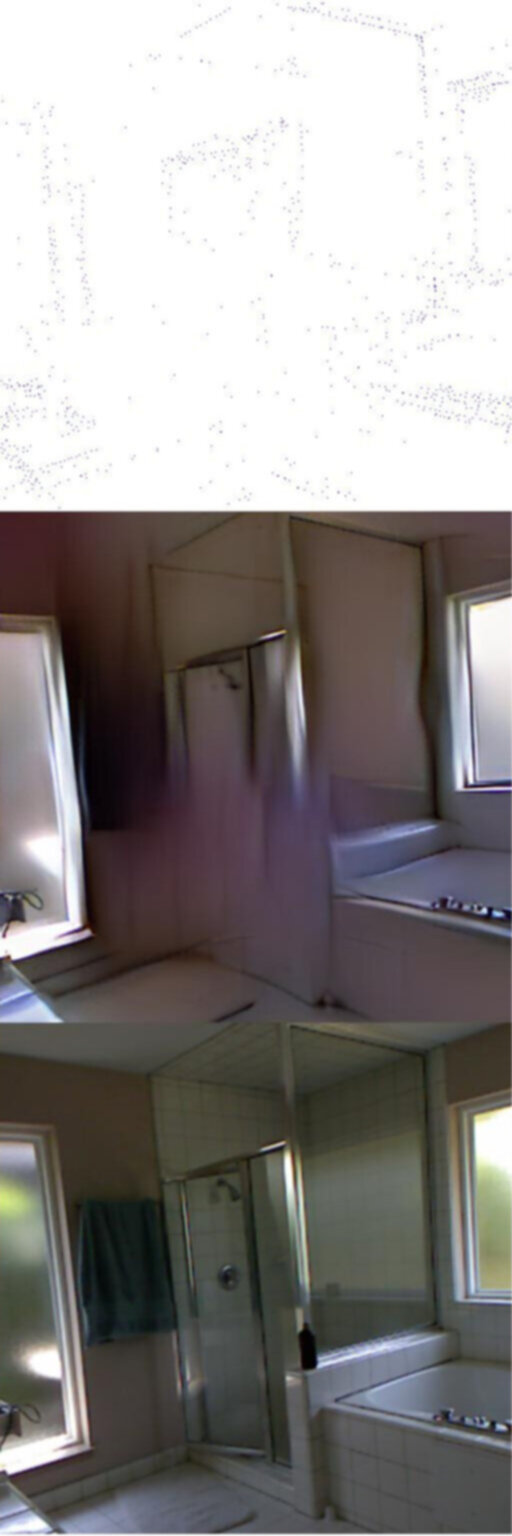}}
\fbox{\includegraphics[width=0.10\linewidth,height=0.3\linewidth]{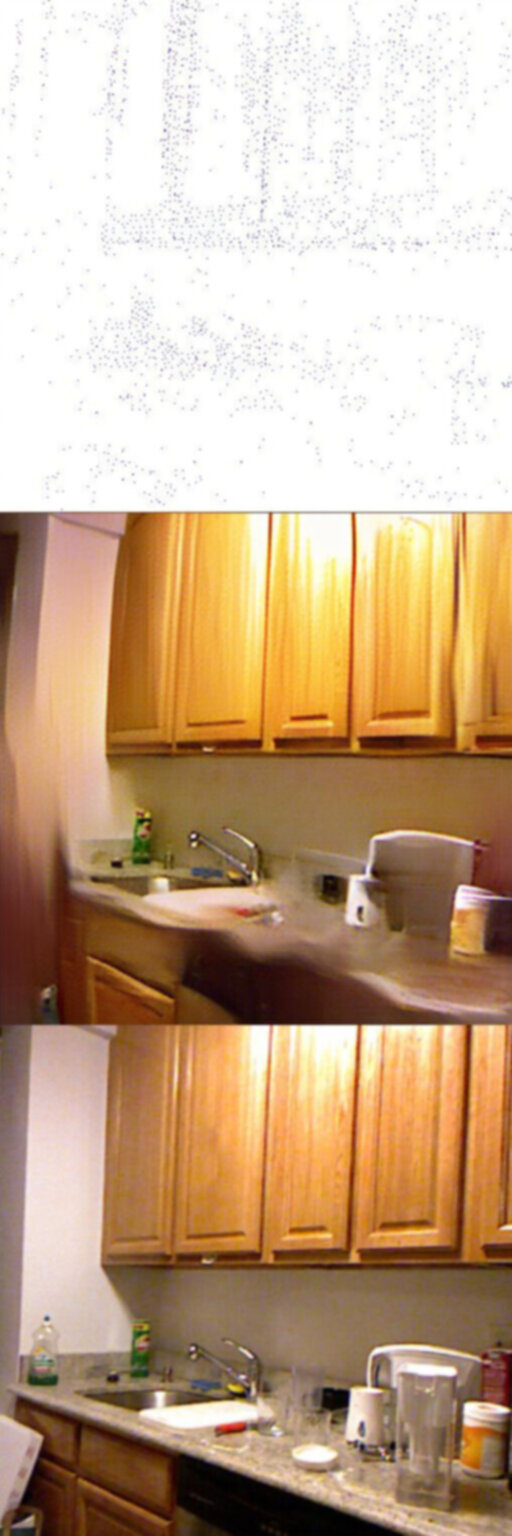}}
\fbox{\includegraphics[width=0.10\linewidth,height=0.3\linewidth]{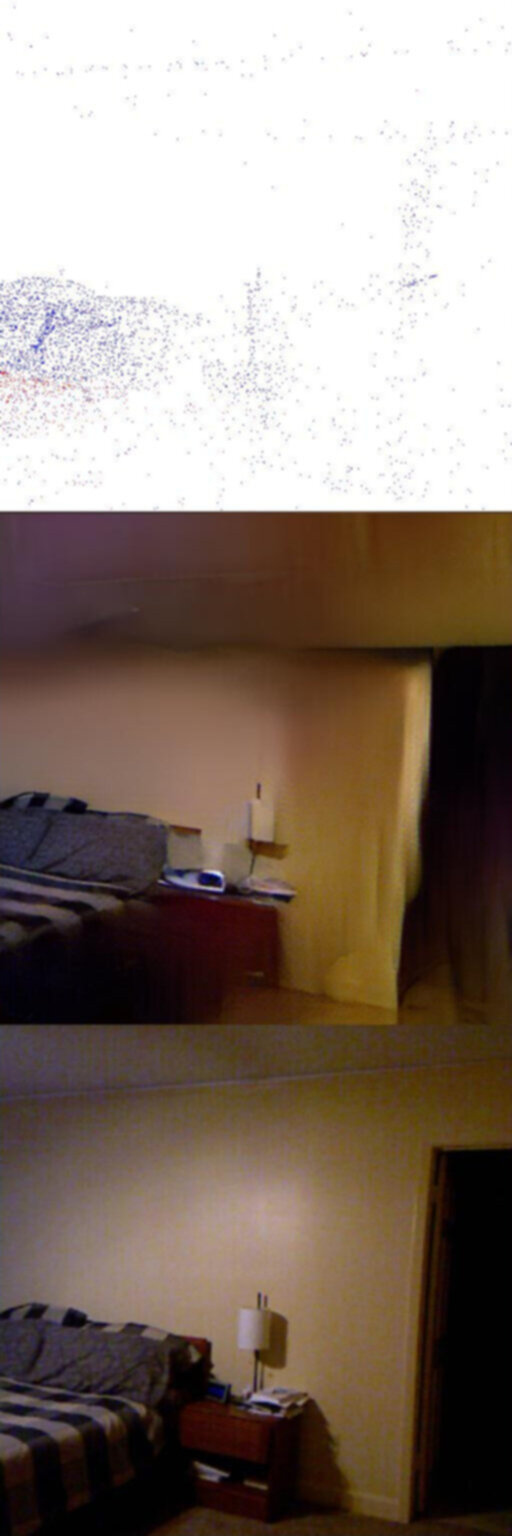}}\\
(a) \hspace{1.5cm} (b) \hspace{1.5cm} (c) \hspace{1.5cm} (d) \hspace{1.5cm} (e) \hspace{1.5cm} (f) \hspace{1.5cm} (g) \hspace{1.5cm} (h)
\caption{\textsc{Failure Examples:} (a) Dense points on the building in the background overwhelms a few sparse points in the foreground on the base of the statue. \textsc{VisibNet} in this case incorrectly predicts that the building is visible and this causes the base of the statue to disappear completely in the synthesized image. (b) A similar artifact for a different scene. (c) Parallel straight lines are sometimes poorly handled, such as the lines on the vertical pillars of the monument. (d) The complex occlusions in the architectural structure produce artifacts where the occluded surfaces and the occluders are fused into each other. (e) Straight lines are often reconstructed as curved or bent (f--g) Low sample density in the input common in indoor scenes results in blurry and wavy edges. (h) Finally, spurious 3D points may cause our method to hallucinate structures such as the dark line on the wall which is not actually there.}
\label{fig:failure}
\end{figure*}

We now present qualitative results to show that our network is robust to 2D input which is very sparse.
Figure \ref{fig:sparsity} shows two example results. Three images are synthesized on randomly selected 20\%, 60\% and 100\%
of all the projected 3D points for the scenes. Despite the high simulated 2D sparsity in the input, the output images
are quite interpretable. Figure~\ref{fig:failure} shows some failure examples.

\vspace{2mm}
\noindent \textbf{Supplementary Video.} Finally, we encourage the reader to view the supplementary video which
makes it easier to visualize the qualitative results shown in the main paper. For two scenes, where
the SfM camera poses are available, we show that we can reconstruct the source video by running our
method on a frame by frame basis with the camera poses for the source images. Finally, we show results of
synthesizing images from novel camera viewpoints. Such results can be used to create virtual tours
of the scene, thus making it easier to reveal and the appearance, layout and geometry of the scene.

\end{document}